\documentclass[10pt,twocolumn,letterpaper]{article}

\usepackage{iccv}
\usepackage{times}
\usepackage{epsfig}
\usepackage{graphicx}
\usepackage{amsmath}
\usepackage{amssymb}
\usepackage{caption}
\usepackage{booktabs}
\usepackage{enumitem}
\usepackage[misc]{ifsym}

\usepackage[pagebackref=true,breaklinks=true,letterpaper=true,colorlinks,bookmarks=false]{hyperref}

\DeclareMathOperator*{\argmin}{argmin}
\iccvfinalcopy 

\def\iccvPaperID{****} 

\begin{document}

\title{StyleGANEX: StyleGAN-Based Manipulation Beyond Cropped Aligned Faces\vspace{-1mm}}

\author{Shuai Yang \hspace{12pt} Liming Jiang  \hspace{12pt}  Ziwei Liu  \hspace{12pt} Chen Change Loy$^{~\textrm{\Letter}}$\\
S-Lab, Nanyang Technological University\\
{\tt\small \{shuai.yang, liming002,  ziwei.liu, ccloy\}@ntu.edu.sg}\vspace{-1mm}
}


\twocolumn[{%
\renewcommand\twocolumn[1][]{#1}%
\maketitle
\vspace{-2.2em}
\begin{center}
\centering
\includegraphics[width=\linewidth]{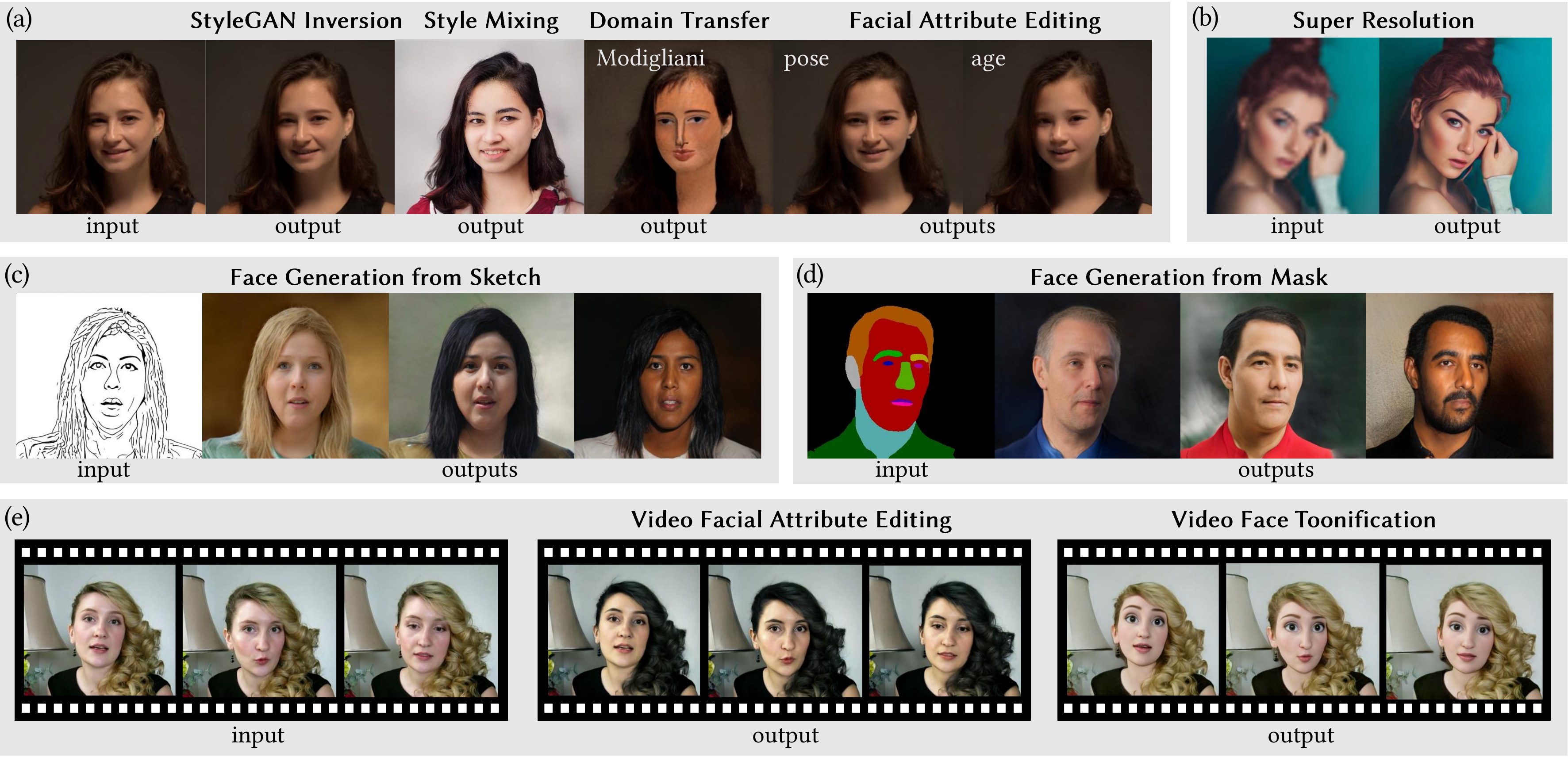}\vspace{-0.3em}
\captionof{figure}{We expand StyleGAN to encompass a diverse set of tasks that go beyond the constraints of cropped aligned faces.}
\label{fig:teaser}
\end{center}%
}]

\begin{abstract}
   Recent advances in face manipulation using StyleGAN have produced impressive results. However, StyleGAN is inherently limited to cropped aligned faces at a fixed image resolution it is pre-trained on.  In this paper, we propose a simple and effective solution to this limitation by using dilated convolutions to rescale the receptive fields of shallow layers in StyleGAN, without altering any model parameters.  This allows fixed-size small features at shallow layers to be extended into larger ones that can accommodate variable resolutions, making them more robust in characterizing unaligned faces. To enable real face inversion and manipulation, we introduce a corresponding encoder that provides the first-layer feature of the extended StyleGAN in addition to the latent style code. We validate the effectiveness of our method using unaligned face inputs of various resolutions in a diverse set of face manipulation tasks, including facial attribute editing, super-resolution, sketch/mask-to-face translation, and face toonification. Project page \url{https://www.mmlab-ntu.com/project/styleganex}.
\end{abstract}

\section{Introduction}
\label{sec:intro}

StyleGAN~\cite{karras2019style,karras2020analyzing} has emerged as one of the most successful models for generating high-quality faces. Building upon StyleGAN, researchers have developed a range of face manipulation models~\cite{abdal2019image2stylegan,richardson2020encoding,tewari2020pie,shen2020interpreting,harkonen2020ganspace,shen2021closed,zhu2021low,wu2021stylealign}. These models typically map real face images or other face-related inputs to the latent space of StyleGAN, perform semantic editing in the latent space, and then map the edited latent code back to the image space. This approach enables a variety of tasks, including facial attribute editing, face restoration, sketch-to-face translation, and face toonification. As the manipulated faces remain within the generative space of StyleGAN, the quality of the image output is guaranteed.

Despite its ability to ensure high-quality image output, the generative space of StyleGAN is limited by a fixed-crop constraint that restricts image resolution and face layout. As a result, existing face manipulation models based on StyleGAN can only handle cropped and aligned face images.
In such images with a limited field of view (FoV), the face typically dominates the image, leaving little room for background and clothing, and often resulting in partially cropped hair. However, in everyday portrait photos such as selfies, faces occupy a smaller proportion of the image, allowing for a complete hairstyle and upper body seen. Portrait videos such as those from live streaming require an even larger background area to accommodate face movement. To process these types of inputs, which we refer to as normal FoV face images and videos, existing manipulation models need to align, crop, and edit the face before pasting the result back onto the original image~\cite{alaluf2022times,liu2022deepfacevideoediting,tzaban2022stitch}. This approach often results in discontinuity near the seams, \eg, only editing the hair color inside the cropped area.

While StyleGAN3~\cite{karras2021alias} was introduced to address unaligned faces, a recent study~\cite{alaluf2022times} found that even StyleGAN3 requires face realignment before effectively projecting to its latent space. Moreover, StyleGAN3 is still constrained by a fixed image resolution.
 Motivated by the translation equivariance of convolutions, VToonify~\cite{yang2022Vtoonify} addresses the fixed-crop limitation of StyleGAN by removing its shallow layers to accept input features of any resolution. However, these shallow layers are crucial for capturing high-level features of the face, such as pose, hairstyle, and face shape. By removing these layers, the network loses its ability to perform latent editing on these important features, which is a distinctive capability of StyleGAN. Therefore, \textit{the challenge remains in overcoming the fixed-crop limitation of StyleGAN while preserving its original style manipulation abilities, which is a valuable research problem to solve}.

In this paper, we propose a simple yet effective approach for refactoring StyleGAN to overcome the fixed-crop limitation. In particular,
we refactor its shallow layers instead of removing them, allowing the first layer to accept input features of any resolution.
This simple change expands StyleGAN's style latent space into a more powerful joint style latent and first-layer feature space ($W^+$\textendash$F$ space), extending the generative space beyond cropped aligned faces.
Furthermore, our refactoring only changes the receptive field of shallow-layer convolutions, leaving all pre-trained model parameters intact. Hence, the refactored StyleGAN (StyleGANEX) can directly load the original StyleGAN parameters,
fully compatible with the generative space of StyleGAN, and retains its style representation and editing ability. This means that the StyleGAN editing vectors found in previous studies~\cite{shen2020interpreting,harkonen2020ganspace,shen2021closed} can be directly applied to StyleGANEX for normal FoV face editing, \eg, changing the face pose, as shown in Fig.~\ref{fig:teaser}(a).

Based on StyleGANEX, we further design a corresponding encoder that projects normal FoV face images to the $W^+$\textendash$F$  space for real face inversion and manipulation.
Our encoder builds upon pSp encoder~\cite{richardson2020encoding} and aggregates its multi-layer features to predict the first-layer feature of StyleGANEX. The encoder and StyleGANEX form a fully convolutional encoder-decoder framework.
With the first-layer feature as the bottleneck layer, whose resolution is $1/32$ of the output image, our framework can handle images and videos of various resolutions, as long as their side lengths are divisible by $32$.
Depending on the input and output types, our framework can perform a wide range of face manipulation tasks. In this paper, we select several representative tasks, as shown in Fig.~\ref{fig:teaser}, including facial attribute editing, face super-resolution, sketch/mask-to-face translation and video face toonification.
While the focus of these tasks in the past is limited to cropped aligned faces, our framework can handle normal FoV faces, showing significant advantages over previous StyleGAN-based approaches.
To summarize, our main contributions are:
\begin{itemize}[itemsep=1.5pt,topsep=2pt,parsep=1.5pt]
  \item A novel StyleGANEX architecture with extended $W^+$\textendash$F$ space , which overcomes the fixed-crop limitation of StyleGAN.
  \item An effective encoder that is able to project normal FoV face images into the $W^+$\textendash$F$ domain.
  \item A generic and versatile fully convolutional framework for face manipulation beyond cropped aligned faces.
\end{itemize}

\section{Related Work}
\label{sec:related_work}

\noindent\textbf{StyleGAN inversion.}
StyleGAN inversion aims at projecting real face images into the latent space of StyleGAN for further manipulation.
Image2StyleGAN~\cite{abdal2019image2stylegan} analyzes the latent space and proposes $W^+$ space to reconstruct real faces with latent code optimization.
PIE~\cite{tewari2020pie} and IDinvert~\cite{zhu2020domain} further consider the editability of the latent code during optimization.
To speed up inversion, pSp~\cite{richardson2020encoding} and e4e~\cite{tov2021designing} train an encoder to directly project the target face to its corresponding latent code, which is however hard to reconstruct fine details and handle occlusions.
To solve this issue, Restyle~\cite{alaluf2021restyle} and HFGI~\cite{wang2022high} predict the residue of latent codes or mid-layer features to reduce errors, respectively.
Instead of focusing on the latent code, PTI~\cite{roich2021pivotal} optimizes StyleGAN itself to fit a target face, which is accelerated by
HyperInverter~\cite{dinh2022hyperinverter} and HyperStyle~\cite{alaluf2022hyperstyle} to predict offsets of network parameters with hyper networks.
The above methods are limited to cropped aligned faces for valid face editing.
With the extended $W^+$ \textendash$F$ space, our framework is able to perform inversion on normal FoV face images.

\noindent\textbf{StyleGAN-based face manipulation.}
An intuitive way of StyleGAN-based face manipulation is to optimize the latent code online to achieve certain objectives such as a pixel-level constraints~\cite{abdal2019image2stylegan}, sketch-based structure constraints~\cite{liu2022deepfacevideoediting} or text-guided semantic constraints~\cite{patashnik2021styleclip,gal2022stylegan}.
Another way is to search for offline editing vectors to add to the latent code for manipulation.
Supervised methods identify meaningful editing vectors based on attribute labels or pre-trained classifiers~\cite{shen2020interpreting,shen2020interfacegan,jiang2021talkedit,abdal2021styleflow}.
On the other hand, unsupervised methods statistically analyze the StyleGAN latent space to discover semantically significant editing directions by principal component analysis~\cite{harkonen2020ganspace}, low-rank factorization~\cite{zhu2021low} and closed-form factorization~\cite{shen2021closed}.
Meanwhile, face manipulation can be directly realized by image-to-image translation frameworks, where StyleGAN is used to generate paired training data~\cite{viazovetskyi2020stylegan2,yang2022Vtoonify} or to build the decoder~\cite{richardson2020encoding,chan2020glean,yang2022Vtoonify}.
To achieve spatial editing, methods~\cite{zhu2021barbershop,parmar2022spatially} are proposed to manipulate mid-layer features in addition to the latent code, similar to our $W^+$\textendash$F$ space.
Moreover, BDInvert~\cite{kang2021gan} and StyleHEAT~\cite{yin2022styleheat} introduces feature transformations for unaligned face editing. However, the above methods, as well as HFGI~\cite{wang2022high}, follow StyleGAN features' original fixed resolution, thus still suffering the crop limitation. Differently, our method predicts the first-layer feature that can have various resolutions in StyleGANEX.
Moreover, as we will show later in Sec.~\ref{sec:compare}, compared to VToonify~\cite{yang2022Vtoonify}, StyleGANEX retains the complete style manipulation abilities of the shallow layers of StyleGAN, with a jointly trained latent code and feature extractor, preserving vivid details and supporting more diverse manipulation tasks.

\section{StyleGANEX}

We begin by briefly analyzing the fixed-crop limitation of StyleGAN, with a particular focus on StyleGAN2~\cite{karras2020analyzing}, since most existing StyleGAN-based manipulation models are built upon it.

\subsection{Analysis of the Fixed-Crop Limitation}
\label{sec:analysis}

\begin{figure}[t]
\centering
\includegraphics[width=\linewidth]{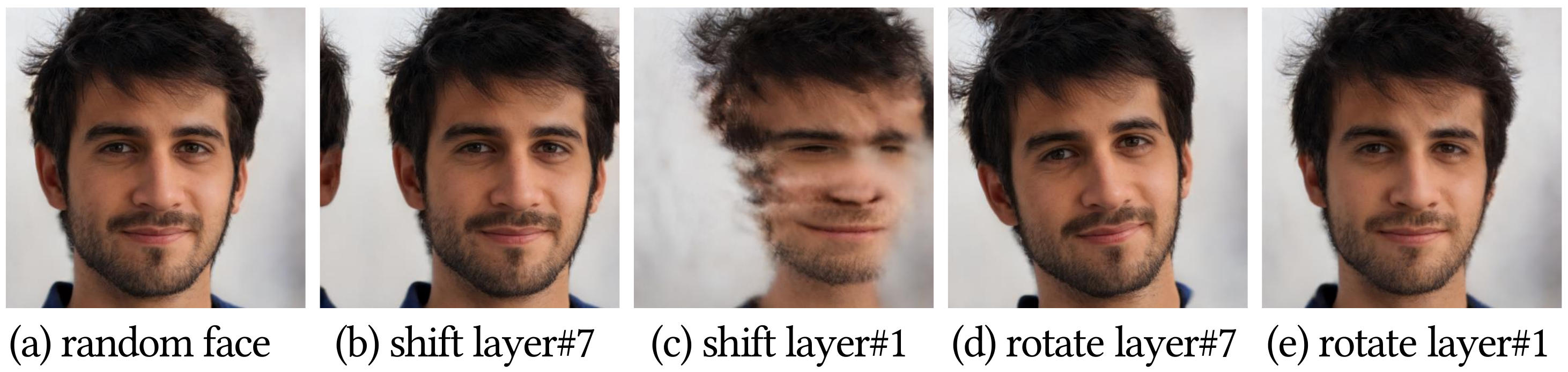}\vspace{-1mm}
\caption{\textbf{Analysis of StyleGAN} in generating unaligned faces. (a) Generated face. (b)(c) Face generated by translating the feature maps of the 7-th and 1-st layers of StyleGAN respectively to shift the face by 150 pixels. (d)(e)  Face generated by rotating the feature maps of the 7-th and 1-st layers of StyleGAN by 10 degrees, respectively.}
\label{fig:analysis1}\vspace{1em}
\centering
\includegraphics[width=\linewidth]{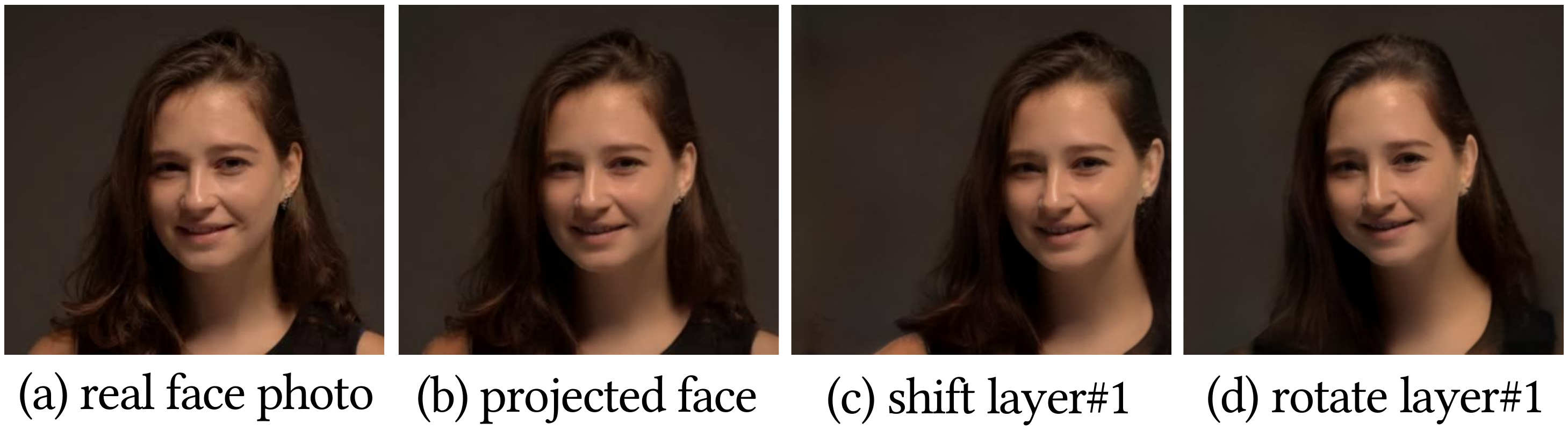}\vspace{-1mm}
\caption{\textbf{Analysis of StyleGANEX} in generating unaligned faces. (a) A real face photo. (b) Reconstructed face by projecting (a) into the $W^+$\textendash$F$  space of StyleGANEX.  (c) Face generated by translating the first-layer feature map to shift the face by 300 pixels. (d) Face generated by rotating the first-layer feature map by 10 degrees.}\vspace{-2mm}
\label{fig:analysis2}
\end{figure}

StyleGAN has great potential for handling normal FoV face images.
The generator of StyleGAN is a fully convolutional architecture that can naturally handle different feature resolutions, as convolution operations and the style modulation of StyleGAN are independent of the input resolution. Additionally, the translation equivariance of convolution operations naturally supports feature translation. As analyzed in VToonify~\cite{yang2022Vtoonify}, if we translate or rotate the feature of the 7th layer of StyleGAN, the resulting face will also be shifted or rotated, as shown in Figs.~\ref{fig:analysis1}(a)(b)(d).

The limiting factor originates from StyleGAN's constant first-layer feature.
First, the first-layer feature has a fixed resolution of $4\times4$, limiting the output to $1024\times1024$ resolution. Second, $4\times4$ resolution is inadequate to characterize the spatial information of unaligned faces.
We have taken a step further than the analysis in VToonify~\cite{yang2022Vtoonify} to investigate translation and rotation on the first-layer feature.
As shown in Fig.~\ref{fig:analysis1}(c), sub-pixel translation fuses adjacent feature values severely, resulting in a blurry face due to the small number of elements (only 16) in the first-layer feature. In Fig.~\ref{fig:analysis1}(e), the first-layer feature fails to provide enough spatial information for a valid rotation. In comparison, the 7-th layer has a higher resolution ($32\times32$), making it better suited for capturing spatial information. However, only a single layer alone provides limited style control, as the full facial structural styles are hierarchically modeled by seven shallow layers. Simply ignoring low-resolution layers, as in VToonify, disables flexible latent editing over the styles of these layers, such as pose, age, and face shape.

Expanding the shallow layers of StyleGAN to have the same $32\times32$ resolution as the 7th layer, or more generally, $H/32\times W/32$ resolution for an $H\times W$ image, would provide enough structure and layout information to combine the style controllability of shallow layers with support for normal FoV faces. This is precisely the key idea of StyleGANEX, and we will introduce our simple solution in Section~\ref{sec:styleganex}.
As a preview of the performance of the expanded layers in enabling face manipulation beyond cropped and aligned faces, Fig.~\ref{fig:analysis2} shows that for a $1472\times1600$ normal FoV face photo, we can obtain its latent code and an additional $46\times50$ first-layer feature as the input to StyleGANEX (inversion method explained in Section~\ref{sec:inversion}). Face translation and rotation can be realized by shifting or rotating the first-layer feature. Additionally, the face can be effectively edited by applying style mixing~\cite{karras2019style} or InterFaceGAN editing vectors~\cite{shen2020interpreting} to the latent code, as shown in Fig.~\ref{fig:teaser}(a).

\subsection{From StyleGAN to StyleGANEX}
\label{sec:styleganex}

\begin{figure}[htbp]
\centering
\includegraphics[width=\linewidth]{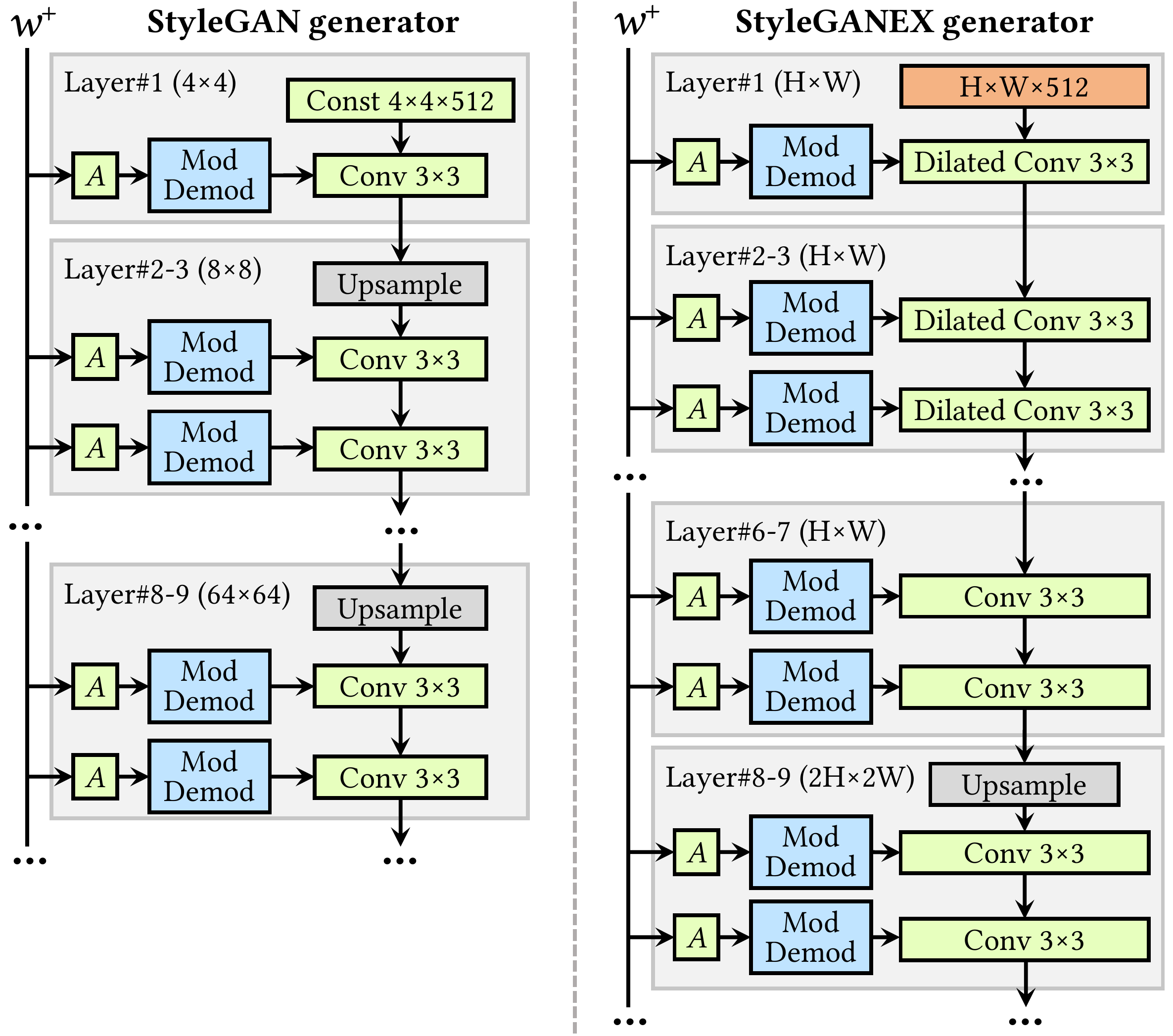}
\caption{\textbf{Refactor StyleGAN to StyleGANEX.} For simplicity, learned weights, biases and noises are omitted.}\vspace{-0.5em}
\label{fig:network}
\end{figure}

Figure~\ref{fig:network} illustrates the generator architectures of StyleGAN and StyleGANEX.
Compared to StyleGAN, we first replace the constant $4\times4$ first-layer feature with a variable feature whose resolution is $1/32$ of the output image.
Then, we remove the upsample operations before the 8-th layer, allowing features in the seven shallow layers to share the same resolution as the 7-th layer.
However, the convolution kernels or reception fields of these layers do not match their input features with the enlarged resolution.
To solve this problem, we enlarge the reception fields by modifying the convolutions to their dilated versions.
For example, the first layer only needs to change the dilation factor from $1$ to $8$.
With this simple modification, StyleGAN is refactored to StyleGANEX.
Since the first layer becomes variable, the original $W^+$ latent space is extended to a joint $W^+$ \textendash$F$ space, where the latent code $w^+\in W^+$ provides style cues, and the first-layer feature $f\in F$ mainly encodes spatial information.

The refactoring of StyleGAN to StyleGANEX has three key advantages.
\textbf{1) Support for unaligned faces.} The resolution enlargement and variable first-layer features of StyleGANEX overcome the fixed-crop limitation.
\textbf{2) Compatibility.} No model parameters are altered during refactoring, meaning that StyleGANEX can directly load pre-trained StyleGAN parameters without retraining. In fact, if we upsample the StyleGAN's constant input feature by $8\times$ with nearest neighbor interpolation to serve as $f$ of StyleGANEX, StyleGANEX degrades exactly to StyleGAN with the same $1024\times1024$ generative space (Fig.~\ref{fig:compatible}). The computational cost of the refactoring is also minimal, with generating an image taking 0.026s and 0.028s for StyleGAN and StyleGANEX, respectively.
\textbf{3) Flexible manipulation.} StyleGANEX retains the style representation and editing ability of StyleGAN, meaning that abundant StyleGAN-based face manipulation techniques can be applied to StyleGANEX.

\begin{figure}[htbp]
\centering
\includegraphics[width=\linewidth]{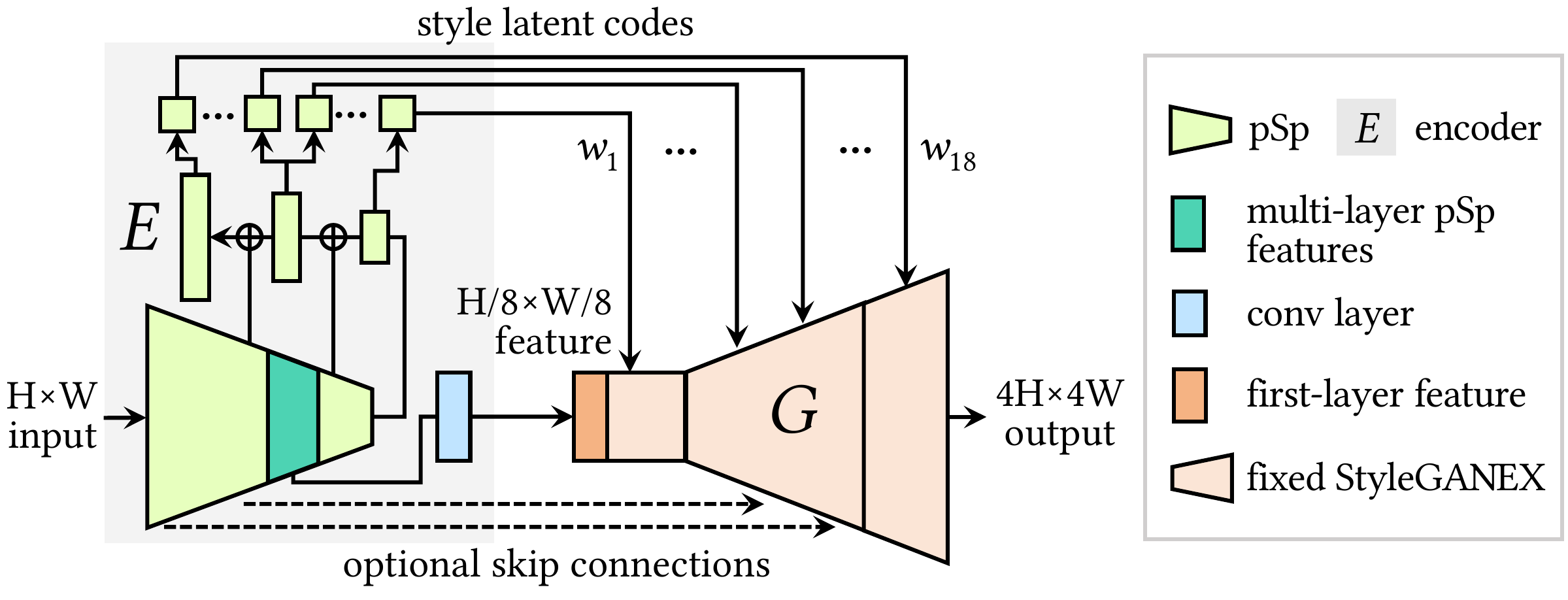}
\caption{\textbf{Details of StyleGANEX Encoder}.}
\label{fig:encoder}\vspace{-1.5em}
\end{figure}

\section{Face Manipulation with StyleGANEX}

\subsection{StyleGANEX Encoder}

This section introduces our StyleGANEX encoder $E$, which is used to project real face images into the $W^+$ \textendash$F$ space of StyleGANEX $G$.
The encoder builds upon the pSp encoder~\cite{richardson2020encoding}, as depicted in Fig.~\ref{fig:encoder}.
Specifically, for $F$ space, we concatenate pSp features in the middle layers and add a convolution layer to map the concatenated features to the first-layer input feature $f$ of $G$.
For $W^+$  space, the original pSp encoder takes a $256\times256$ image as input and convolves it to eighteen $1\times1\times512$ features to map to a latent code $w^+\in \mathbb{R}^{18\times512}$. To make $E$ accept more general $H\times W$ images, we add global average pooling to resize all features to $1\times1\times512$ before mapping to latent codes.
To support various face manipulation tasks flexibly, we can extract $f$ and $w^+$ from different sources.
Let $E_F$ and $E_W$ be the operation of $E$ to extract the first-layer feature and the latent code, respectively.
We have
\begin{equation}
\label{eq:encoder}
  f, w^+= E_F(x_1), E_W(x_2) := E(x_1,x_2),
\end{equation}
where $x_1$ and $x_2$ are the source inputs for face layout and face style, respectively. Then, a general form of image generation by $G$ from $x_1$ and $x_2$ is $G(E_F(x_1), E_W(x_2))$.
In some face manipulation tasks like super-resolution~\cite{chan2020glean} and toonification~\cite{yang2022Vtoonify}, passing encoder features to the generator via skip connections helps preserve the details of the input image. We thus introduce a scalar parameter $\ell$ to the generation process, indicating the $\ell$ shallow layers of $G$ receive the encoder features ($\ell=0$ means no skip connections):
\begin{equation}
\label{eq:generate}
  \hat{x}= G(E_F(x_1, \ell), E_W(x_2)) := G(E(x_1, x_2, \ell)),
\end{equation}
where $E_F(x_1, \ell)$ provides both $f$ and the skipped encoder features.
For $x_1$ of $H\times W$  resolution, $f$ and the generated image $\hat{x}$ will be of $H/8\times W/8$ and $4H\times4W$, respectively.
The resolution of $x_2$ can be independent of $x_1$ and $\hat{x}$.

\subsection{StyleGANEX Inversion and Editing}
\label{sec:inversion}

\begin{figure}[htbp]
\centering
\includegraphics[width=\linewidth]{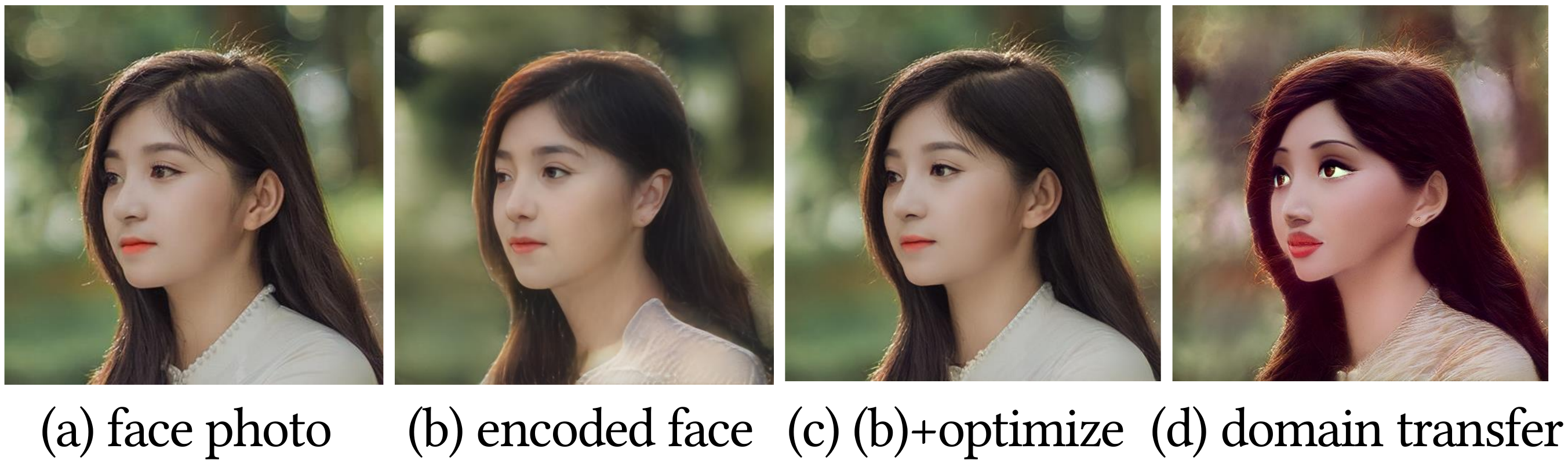}\vspace{-2mm}
\caption{\textbf{StyleGANEX inversion}.}
\label{fig:inversion}\vspace{-1em}
\end{figure}

To find appropriate $\hat{f}$ and $\hat{w}^+$ that precisely reconstruct a target image $x$, we perform a two-step StyleGANEX inversion.
Step I projects $x$ to initial $f$ and $w^+$ with $E$. Step II optimizes $f$ and $w^+$ to further reduce the reconstruction error. The training of $E$ follows pSp with reconstruction losses and a regularization loss~\cite{richardson2020encoding}:
\begin{equation}
\label{eq:inversion}
  \mathcal{L} = \mathcal{L}_\text{rec}(\hat{x}, x)+\lambda_1\mathcal{L}_\text{reg}(E_W({\tilde{x}})),
\end{equation}
where $\hat{x}=G(E(x, \tilde{x}, 0))$ and $\tilde{x}$ is the cropped aligned face region of $x$.
We empirically find that using $\tilde{x}$ instead of $x$ predicts more accurate $w^+$ since StyleGAN is originally trained on cropped aligned faces.
$\mathcal{L}_\text{reg}$ encourages the predicted $w^+$ closer to the average latent code to improve the image quality.
$\mathcal{L}_\text{rec}$ measures the distance between the reconstructed $\hat{x}$ and the target $x$
in terms of pixel similarity, perceptual similarity, and identity preservation:
\begin{equation}
  \mathcal{L}_\text{rec}(\hat{x}, x) = \lambda_2\mathcal{L}_2(\hat{x}, x)+\lambda_3\mathcal{L}_\text{LPIPS}(\hat{x}, x)
  +\lambda_4\mathcal{L}_\text{ID}(\hat{x}, x).\nonumber
\end{equation}
As shown in Fig.~\ref{fig:inversion}(b), $\hat{x}$ largely approximates $x$. But the background details and clothings are still hard to reconstruct. Therefore, we further optimize $f$ and $w^+$
\begin{equation}
  \hat{f}, \hat{w}^+=\argmin_{f, w^+}\mathcal{L}_\text{LPIPS}(G(f, w^+), x),
\end{equation}
where $f$ and $w^+$ are initialized by $E(x, \tilde{x}, 0)$. The optimized $G(\hat{f}, \hat{w}^+)$  in Fig.~\ref{fig:inversion}(c) well reconstructs $x$.

After inversion, we can perform flexible editing over $x$ as in StyleGAN. Figure~\ref{fig:teaser}(a) shows two examples: we can exchange the last 11 elements of $\hat{w}^+$ with random samples, to mix the color and texture styles; we can add InterFaceGAN editing vectors~\cite{shen2020interpreting} to $\hat{w}^+$ to make a young face. Moreover, as shown in Fig.~\ref{fig:inversion}(d),
if we load $G$  a pre-trained StyleGAN-NADA Disney Princess model~\cite{gal2022stylegan} (let $G'$ denote the new $G$), we can obtain $G'(\hat{f}, \hat{w}^+)$, a cartoon version of $x$.

\subsection{StyleGANEX-Based Translation}\label{sec:translation}

The encoder and StyleGANEX form an end-to-end image-to-image translation framework in Fig.~\ref{fig:encoder}.
Depending on the type of paired training data, it can be trained to efficiently realize different face manipulation tasks. As with pSp~\cite{richardson2020encoding}, we will fix StyleGANEX generator and only train the encoder on the given task.

\textbf{Face super-resolution}. Given low-resolution and high-resolution training image pairs $(x,y)$, we can train $E$ to recover $y$ from $x$ to learn face super-resolution with the loss
\begin{equation}
  \mathcal{L} = \mathcal{L}_\text{rec}(\hat{y}, y)+\lambda_5\mathcal{L}_\text{adv}(\hat{y}, y),
\end{equation}
where $\hat{y}=G(E(x_\uparrow, \tilde{x}_\uparrow, 7))$, $\uparrow$ is the upsample operation to make $x$ match the input resolution of $E$, $\tilde{x}$ is the cropped aligned face region of $x$. We add an adversarial loss $\mathcal{L}_\text{adv}$ to improve the realism of the generated image.

\textbf{Sketch/mask-to-face translation}. Given a real face $y$ as target and its sketch or parsing mask $x$ as source, we can train $E$ to translate $x$ to $y$ with Eq.~(\ref{eq:inversion}) as objectives. In this task, we add a trainable light-weight translation network $T$ to map $x$ to an intermediate domain where $E$ can more easily extract features. For the style condition, $G$'s first 7 layers use the latent code extracted from $x$ to provide structural styles, while its last 11 layers use the latent code from $\tilde{y}$ to provide color and texture styles to simplify reconstruction. Therefore, $\hat{y}=G(E_F(T(x), \ell), E^{1:7}_W(T(x))\oplus E^{8:18}_W(\tilde{y}))$, where $\oplus$ is concatenation operation, and the superscript of $E_W$ means taking the 1$\sim$7 or 8$\sim$18 elements of $w^+$. $\ell=1$ for sketch inputs and $\ell=3$ for mask inputs.

\textbf{Video face editing}. Given paired original face,  edited face, and its editing vector $(x,y,v)$ (we can simply generate $x=G_0(w^+)$ and $y=G_0(w^++v)$  from random latent code $w^+$ with StyleGAN $G_0$), we train $E$ for face editing with
\begin{equation}
\label{eq:video}
  \mathcal{L} = \mathcal{L}_\text{rec}(\hat{y}, y)+\lambda_5\mathcal{L}_\text{adv}(\hat{y}, y)+\lambda_6\mathcal{L}_\text{tmp}(\hat{y}),
\end{equation}
where $\hat{y}=G(E_F(x, 13), E_W(\tilde{x})+v)$. $\mathcal{L}_\text{tmp}$ is the flicker suppression loss~\cite{yang2022Vtoonify} to improve temporal consistency.

\textbf{Video toonification}. For video face toonification, we have paired original face and toonified face $(x,y)$.
They can be generated as $x=G_0(w^+)$ and $y=G'_0(w^+)$ from random latent code $w^+$ following Toonify~\cite{pinkney2020resolution}, where $G'_0$ is the StyleGAN fine-tuned on cartoon images. Let $G'$ denote StyleGANEX loaded with $G'_0$, then we train $E$ using the objectives of Eq.~(\ref{eq:video}) with $\hat{y}=G'(E(x, \tilde{x}, 13))$.

Note that compared to face editing based on StyleGANEX inversion in Sec.~\ref{sec:inversion}, the solution in this section does not require time-consuming latent optimization, and temporal consistency is enforced by the flicker suppression loss. Therefore, this solution is more suitable for efficient and coherent video face manipulation, which is a unique feature of the proposed framework.
On the other hand, the solution based on StyleGANEX inversion is more flexible. There is no need to train new $E$ for every editing vector $v$ or fine-funed StyleGAN $G'_0$.

\begin{figure*}[t]
\centering
\includegraphics[width=\linewidth]{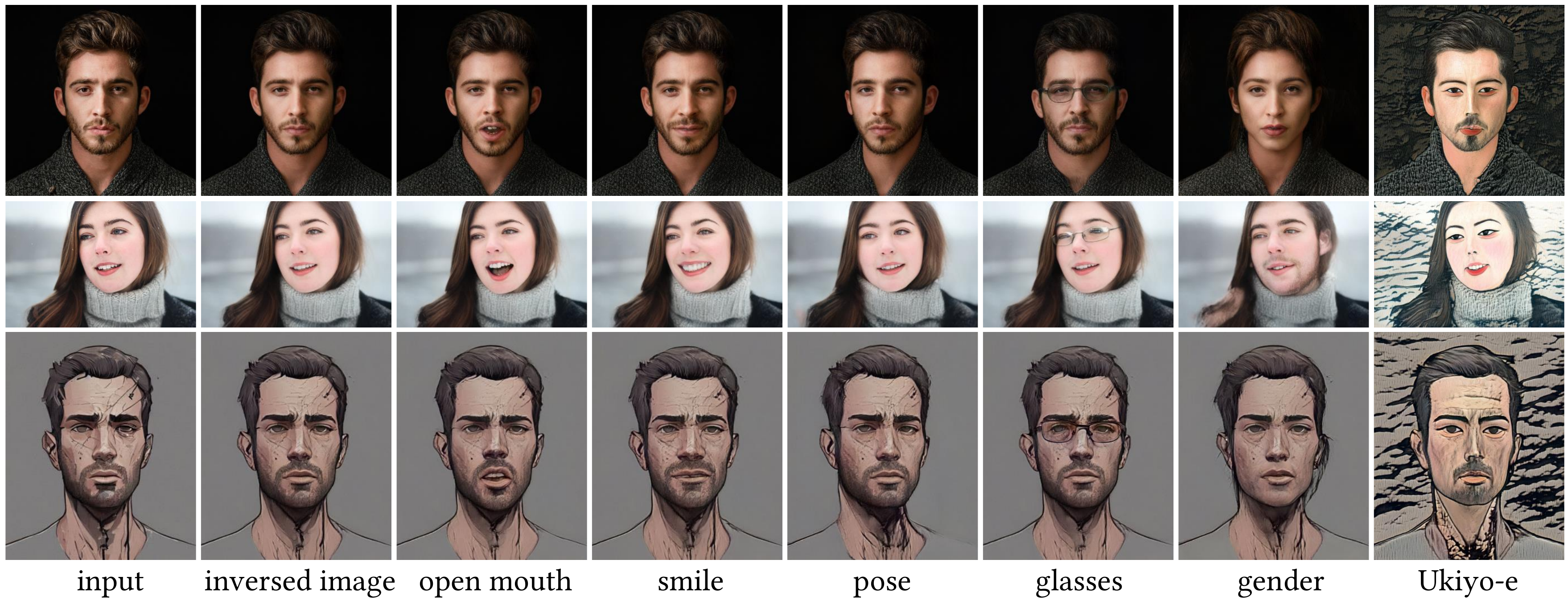}\vspace{-2mm}
\caption{\textbf{StyleGANEX inversion and facial attribute/style editing}.}
\label{fig:encode}\vspace{-1.2em}
\end{figure*}

\section{Experimental Results}

\noindent
\textbf{Implementation details}. We follow pSp~\cite{richardson2020encoding} to set $\lambda_2=1$ and $\lambda_3=0.8$ for all tasks, $\lambda_4=0.1$ for inversion task and $0$ for other tasks.
We set $\lambda_1$ to $0.0001$, $0.005$ and $0$ for inversion, sketch/mask-to-face and other tasks, respectively. We set $\lambda_5=0.1$ and $\lambda_6=30$ empirically.  The translation network $T$ consists of two downsampling convolutional layers, two ResBlocks~\cite{he2016deep} and two upsampling convolutional layers, with small channel number $16$. All experiments are performed using a single NVIDIA Tesla V100 GPU.

\noindent
\textbf{Datasets.}~We process FFHQ~\cite{karras2019style} to obtain 70,000 aligned training images of $1280\times1280$  resolution for all tasks except two video-related tasks that use StyleGAN generated data. We use BiSeNet~\cite{Yu-ECCV-BiSeNet-2018} to extract parsing masks and follow pSp~\cite{richardson2020encoding} to extract sketches from face images.
We augment all training data with random geometric transformations~\cite{karras2020training} like scaling, translation and rotation to make faces unaligned.
We use images and videos from FaceForensics++~\cite{roessler2019faceforensicspp}, Unsplash and Pexels as our testing dataset.


\subsection{Face Manipulation}\label{sec:compare}

\begin{figure}[t]
\centering
\includegraphics[width=\linewidth]{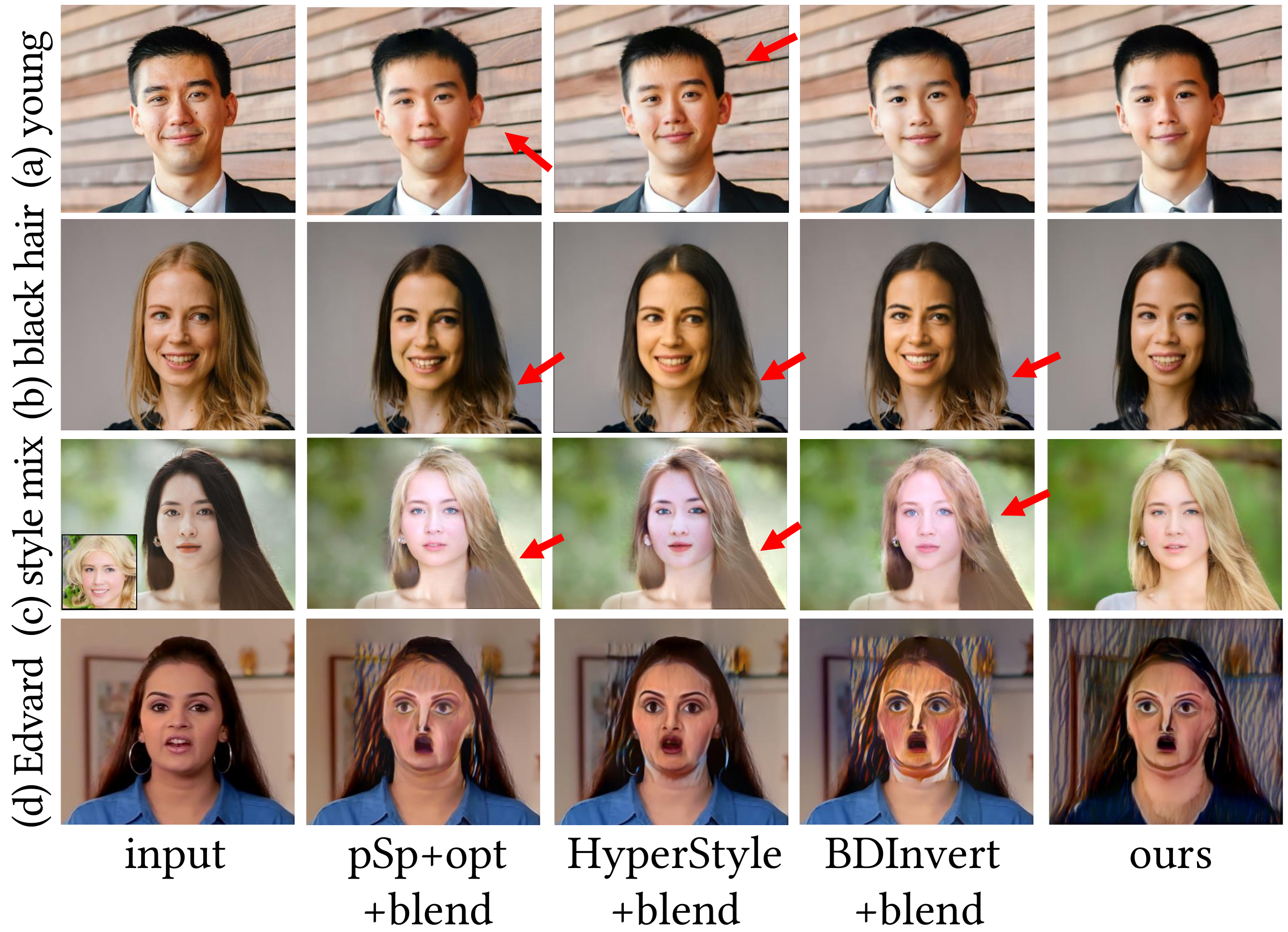}\vspace{-2mm}
\caption{\textbf{Comparison on face editing}. The local region in the yellow box is shown enlarged below each image.}
\label{fig:editing}\vspace{1mm}
\includegraphics[width=\linewidth]{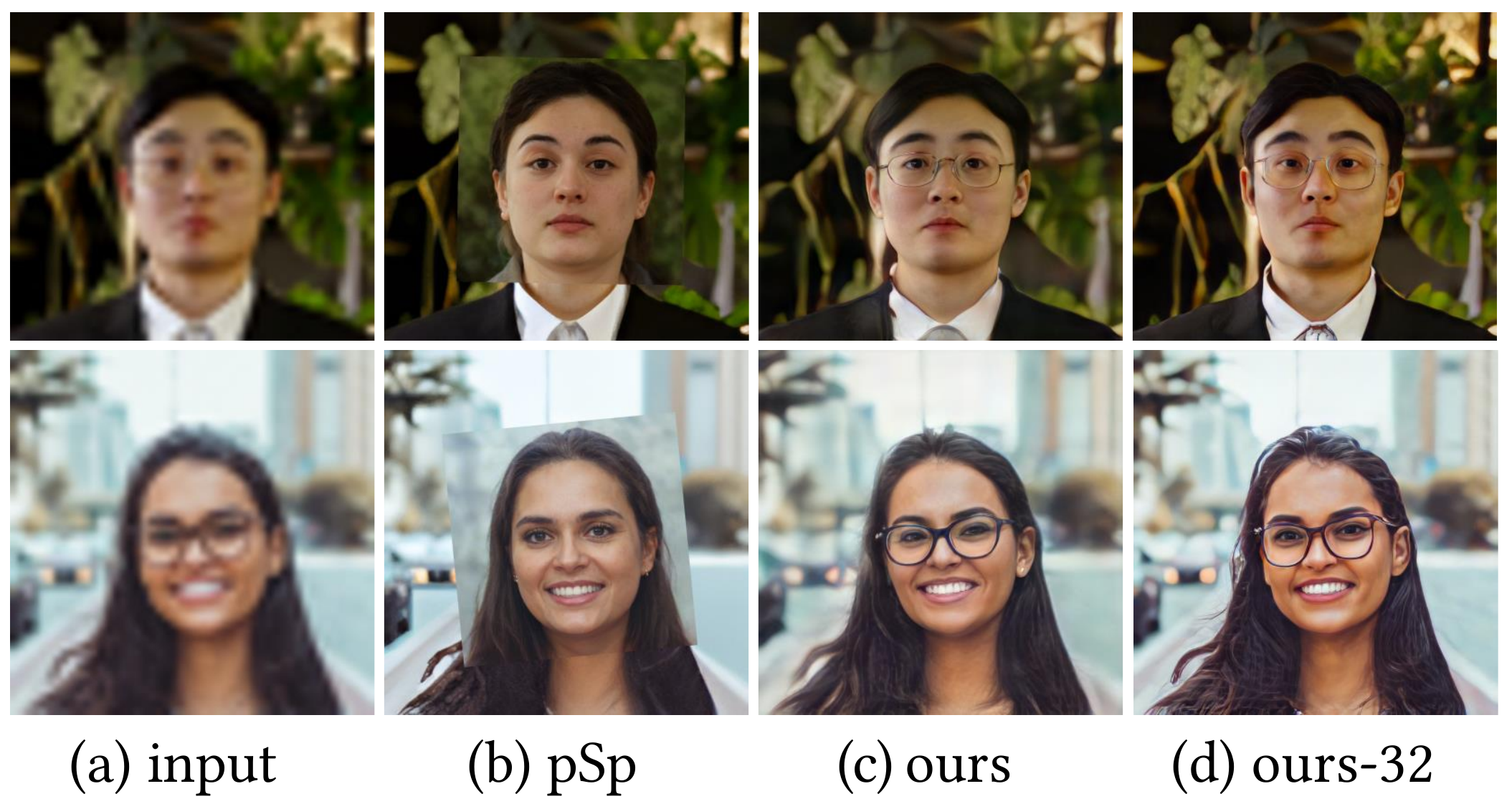}\vspace{-2mm}
\caption{\textbf{Comparison on super-resolution}.}
\label{fig:sr}\vspace{-1.4em}
\end{figure}

\noindent
\textbf{Face editing}. Figure~\ref{fig:encode} provides an overview of the performance of face inversion and attribute editing on StyleGANEX. We apply inversion to normal FoV face photos/paintings and use various editing vectors from InterFaceGAN~\cite{shen2020interpreting} and LowRankGAN~\cite{zhu2021low}, and the pre-trained StyleGAN-NADA Ukiyo-e model~\cite{gal2022stylegan}, to edit the facial attributes or styles. As shown, these StyleGAN editing techniques work well on StyleGANEX.
We also compare with pSp~\cite{richardson2020encoding}, HyperStyle~\cite{alaluf2022hyperstyle} and BDInvert~\cite{kang2021gan} in Fig.~\ref{fig:editing}. Since these baselines are designed for cropped faces, we paste and blend their edited results back into the original image. For a fair comparison, we apply the same optimization method used in our approach to pSp for precise inversion (HyperStyle already uses extra hyper networks to simulate optimization and BDInvert is optimization-based inversion). For editing that alters structures or colors, even precise inversion and blending cannot eliminate the obvious discontinuity along the seams, as indicated by the red arrows. In contrast, our approach processes the entire image as a whole and avoids such issues.
Remarkably, our method successfully turns the whole hair into black in Fig.\ref{fig:editing}(b), transfers the exemplar blond hairstyle onto the target face in Fig.\ref{fig:editing}(c), and renders the full background with the StyleGAN-NADA Edvard Munch style in Fig.\ref{fig:editing}(d).

\noindent
\textbf{Face super resolution}. We show our $32\times$ super-resolution results in Fig.~\ref{fig:sr}(d), where both the face and non-face regions are reasonably restored. We further follow pSp to train a single mode on multiple rescaling factors ($4\sim64$) with $\ell=3$ to make a fair comparison. In pSp's results, the non-face region is super-resolved by
Real-ESRGAN~\cite{wang2021real}. As in Fig.~\ref{fig:sr}(b)(c), our method surpasses pSp in precise detail restoration (\eg, glasses) and uniform super-resolution without discontinuity between face and non-face regions.

\begin{figure}[t]
\centering
\includegraphics[width=\linewidth]{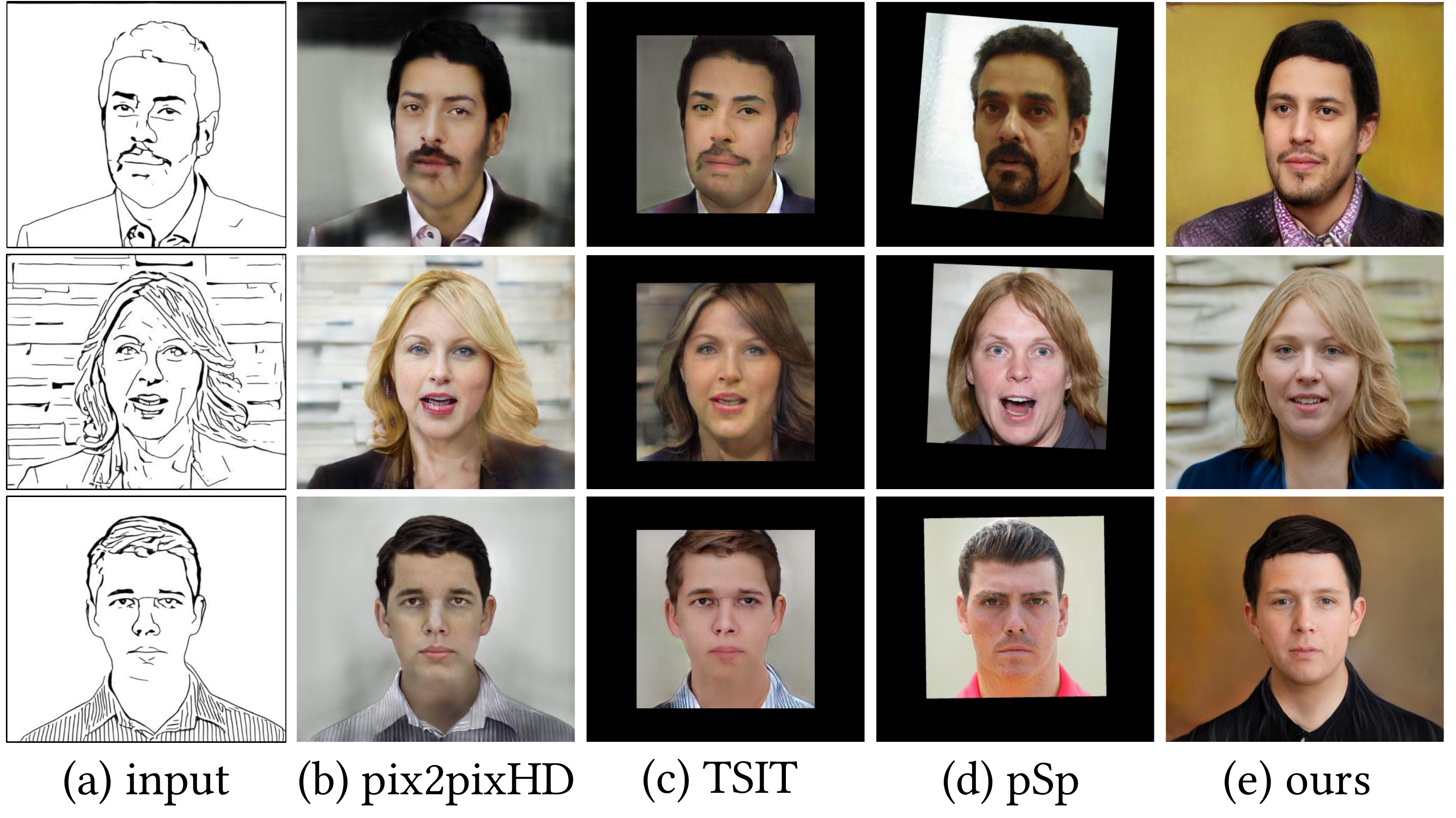}\vspace{-2mm}
\caption{\textbf{Comparison on sketch-to-face translation}.}
\label{fig:sketch2face}\vspace{1mm}
\includegraphics[width=\linewidth]{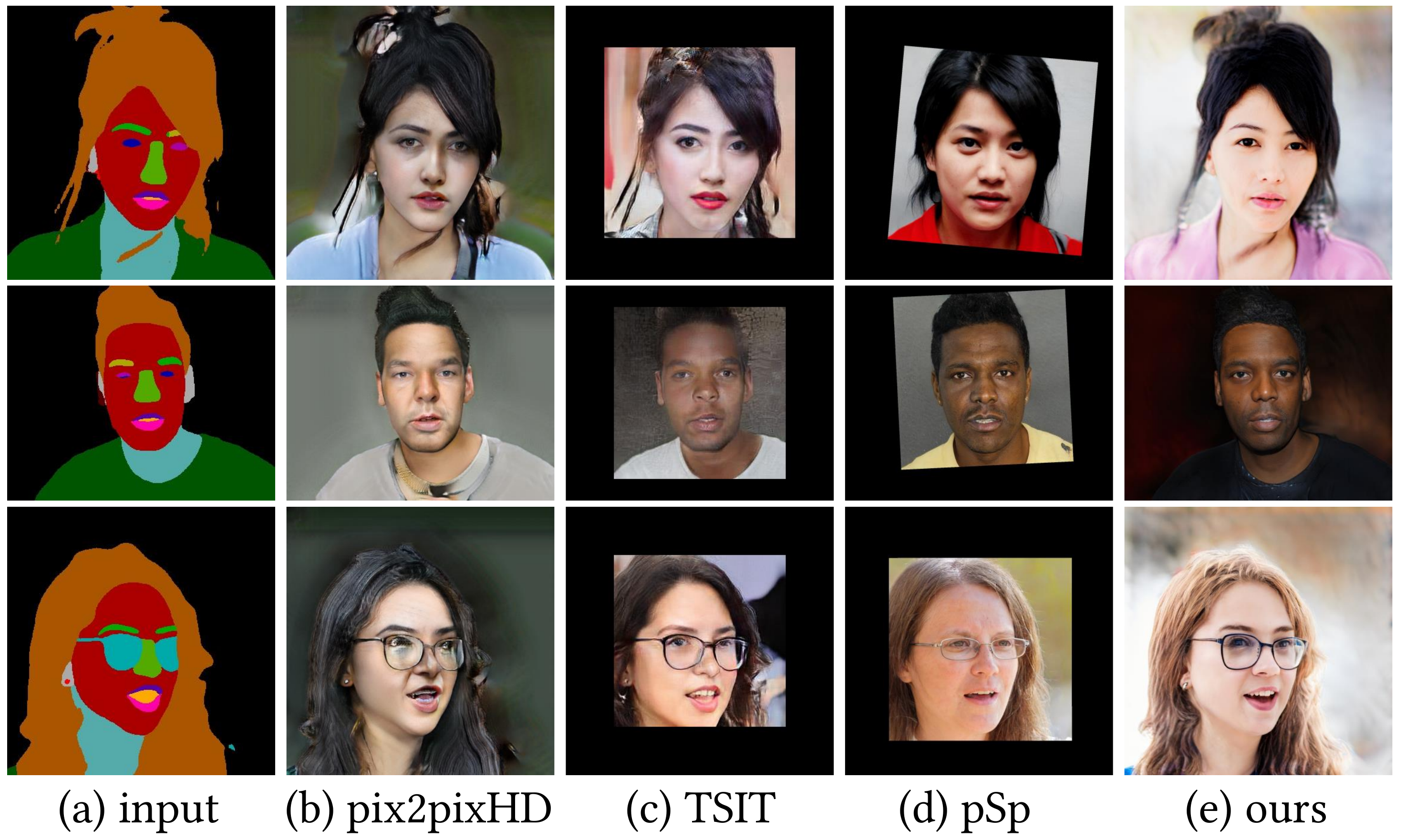}\vspace{-2mm}
\caption{\textbf{Comparison on mask-to-face translation}.}
\label{fig:seg2face}\vspace{-1.2em}
\end{figure}

\begin{figure}[t]
\centering
\includegraphics[width=\linewidth]{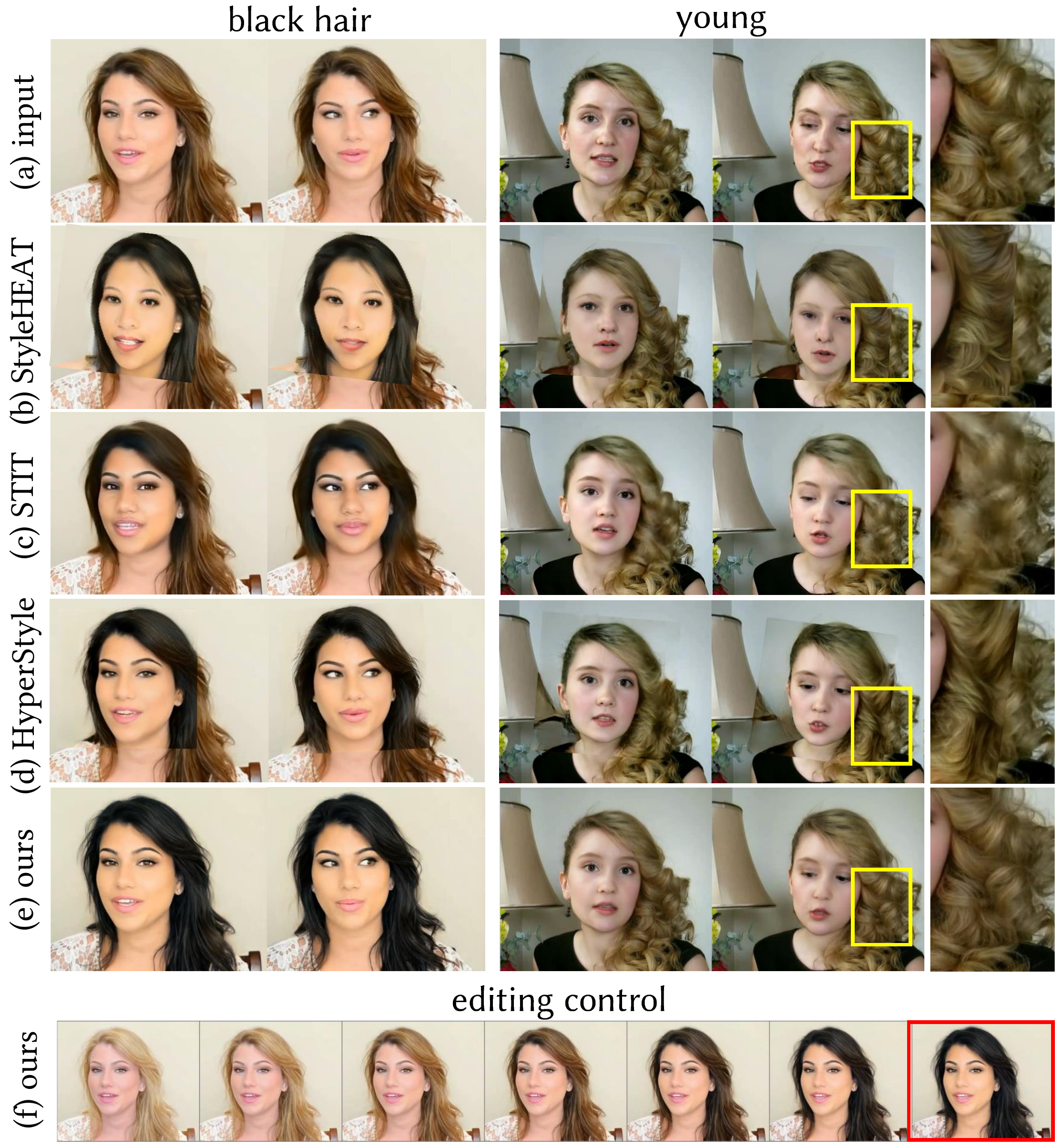}\vspace{-2mm}
\caption{\textbf{Comparison on video face editing}.}
\label{fig:video_editing}\vspace{2mm}
\includegraphics[width=\linewidth]{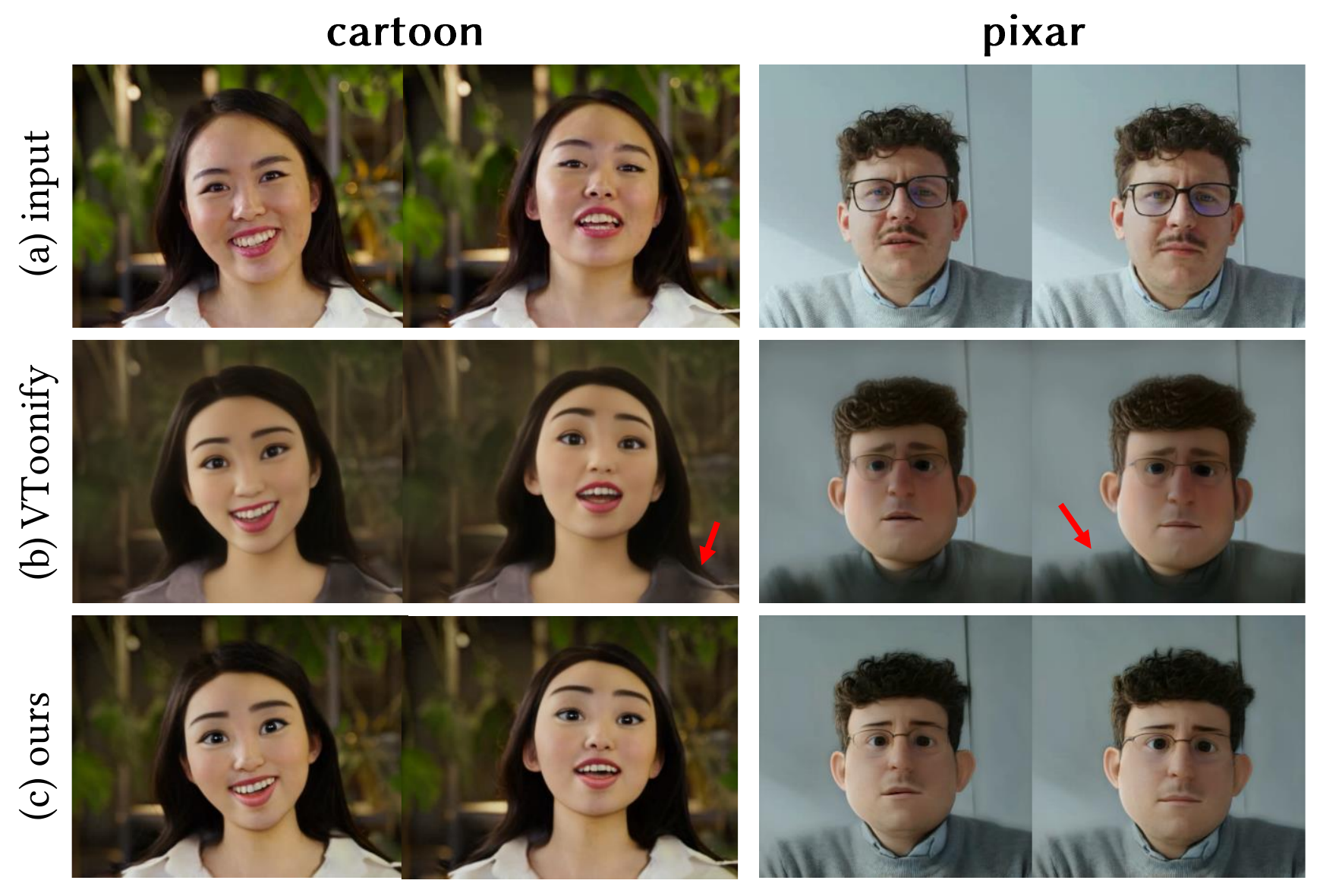}\vspace{-2mm}
\caption{\textbf{Comparison on video face toonification}.}
\label{fig:toonify}\vspace{-1.2em}
\end{figure}

\noindent
\textbf{Sketch/mask to face translation}. We compare our method with image-to-image translation models pix2pixHD~\cite{Wang2017High} and TSIT~\cite{jiang2020tsit}, and StyleGAN-based pSp in Figs.~\ref{fig:sketch2face}-\ref{fig:seg2face}. Pix2pixHD's results have many artifacts and monotonous colors. TSIT requires the inputs' side lengths to be divisible by 128. We find padding the input leads to failed translation. Therefore, we show its results on centrally cropped inputs, which are blurry. PSp generates realistic results, which are however less similar to the input sketch/mask. By comparison, our method can translate whole images and achieve realism and structural consistency to the inputs.
For quantitative evaluation, we conduct a user study, where 30 subjects are invited to select what they consider to be the best results from the four methods. Each task uses eight results for evaluation.
Table~\ref{tb:user_study}  summarizes the preference scores, where our method receives the best score.

\noindent
\textbf{Video face editing}.
We compare with HyperStyle, StyleHEAT~\cite{yin2022styleheat} and STIT~\cite{tzaban2022stitch}. StyleHEAT generates unaligned face videos based on warping the first frame's features, which however limits its inversion accuracy. STIT extends PTI~\cite{roich2021pivotal} for full video processing by stitching. STIT cannot well preserve the complex hair details (Fig.~\ref{fig:video_editing}, yellow box). As with image face editing, all three baselines are limited to editing cropped regions, leading to discontinuity along the stitching seams.
By comparison, our method uses the first-layer feature and skipped mid-layer features to provide spatial information, which achieves more coherent results. Moreover, we can randomly scale the editing vector $v$ (by multiplying a scale factor) instead of using a fixed $v$ during training. Then during testing, our method can flexibly adjust the editing degree by scaling $v$ for users to select as in Fig.~\ref{fig:video_editing}(e).

\noindent
\textbf{Video face toonification}. Compared with VToonify-T~\cite{yang2022Vtoonify}, our method preserves more details of the non-face region and generates shaper faces in Fig.~\ref{fig:toonify}. The reason is that VToonify-T uses a fixed latent code extractor while our method trains a joint latent code and feature extractor, thus our method is more powerful for reconstructing the details. Moreover, our method retains StyleGAN's shallow layers, which helps provide key facial features to make the stylized face more vivid. Table~\ref{tb:user_study} shows a quantitative comparison on ten results, and our method obtains the best user score.

\begin{table} [t]
\caption{\textbf{User preference scores}. Best scores are in bold.}\vspace{-2mm}
\label{tb:user_study}
\centering
\footnotesize
\begin{tabular}{l|c|c|c|c}
\toprule
Task & pix2pixHD~\cite{Wang2017High} & TSIT~\cite{jiang2020tsit} & pSp~\cite{richardson2020encoding} & ours \\
\midrule
sketch-to-face & 0.400 & 0.121 & 0.029 & \textbf{0.450} \\
mask-to-face & 0.108 & 0.075 & 0.121 & \textbf{0.696} \\
\toprule
\toprule
Task & \multicolumn{2}{c|}{VToonify-T~\cite{yang2022Vtoonify}} & \multicolumn{2}{c}{ours} \\
\midrule
video toonify & \multicolumn{2}{c|}{0.083} & \multicolumn{2}{c}{\textbf{0.917}} \\
\bottomrule
\end{tabular}\vspace{-1em}
\end{table}

\subsection{Ablation Study}

\begin{figure}[t]
\centering
\includegraphics[width=\linewidth]{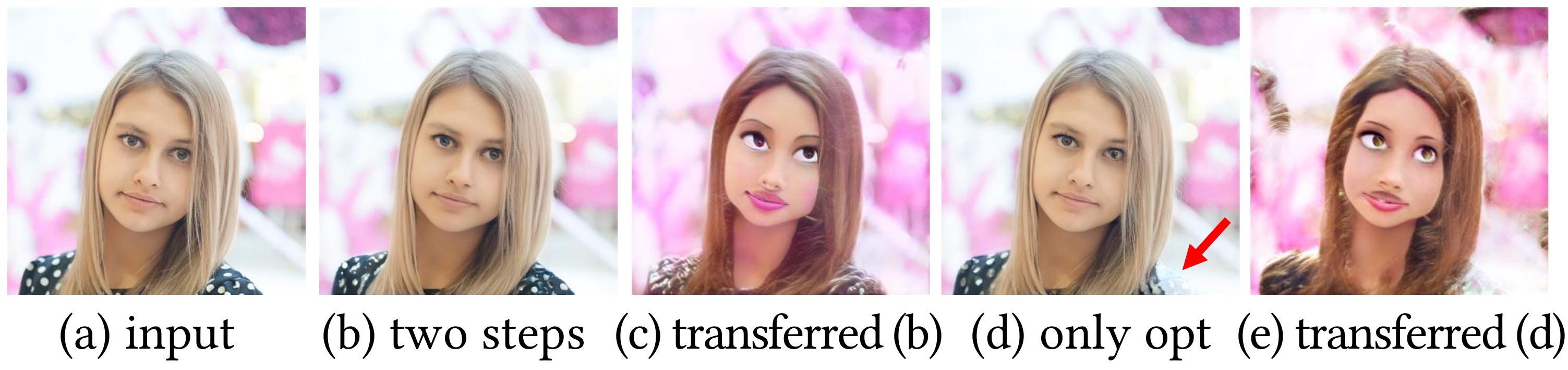}\vspace{-2mm}
\caption{\textbf{Effect of encoder} in StyleGANEX inversion.}
\label{fig:abla_inversion}\vspace{1mm}
\includegraphics[width=\linewidth]{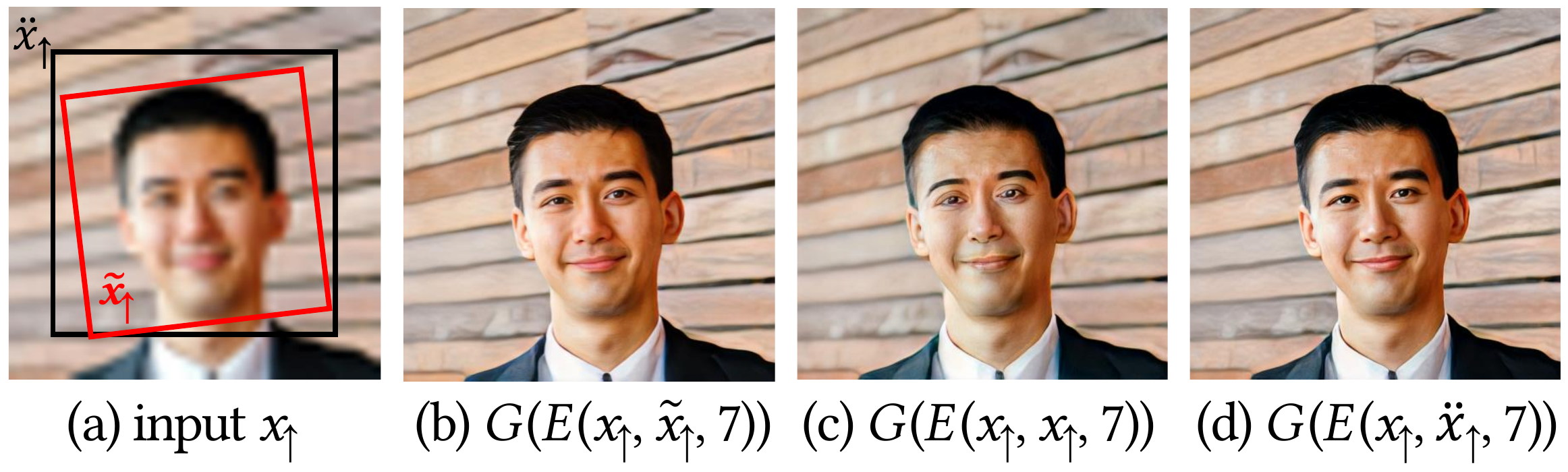}\vspace{-2mm}
\caption{\textbf{Input choice} to provide valid style information.}
\label{fig:abla_crop}\vspace{-1.2em}
\end{figure}

The effect of Step II of our two-step inversion is verified in Fig.~\ref{fig:inversion}. We further study the effect of Step I in Fig.~\ref{fig:abla_inversion}. With Step I providing a good prediction of $w^+$ and $f$, Step II only needs 500-iteration optimization for precise reconstruction (Fig.~\ref{fig:abla_inversion}(b)) and valid domain transfer to Disney Princess (Fig.~\ref{fig:abla_inversion}(c)). However, if we directly optimize a mean $w^+$ and a random $f$, the result is poor even with 2,000 iterations (indicated by a red arrow in Fig.~\ref{fig:abla_inversion}(d)) and the optimized $w^+$ and $f$ overfit the input, which is not suitable for editing like domain transfer (Fig.~\ref{fig:abla_inversion}(e)).

In Fig.~\ref{fig:abla_crop}, we study the input choice to extract $w^+$. The cropped aligned faces are the default choice. If we instead use the whole image to extract $w^+$, the background will affect the facial style, leading to poor restoration in Fig.~\ref{fig:abla_crop}(c). We further find reasonable results (Fig.~\ref{fig:abla_crop}(d)) can still be obtained by cropping the input to decrease the background proportion. Note that the face is not aligned in the cropped image ($\ddot{x}_{\uparrow}$), which is useful for applications like super-resolution where extremely low-resolution faces are hard to detect and align. Users can simply manually crop the face region to provide valid style information.

\begin{figure}[t]
\centering
\includegraphics[width=\linewidth]{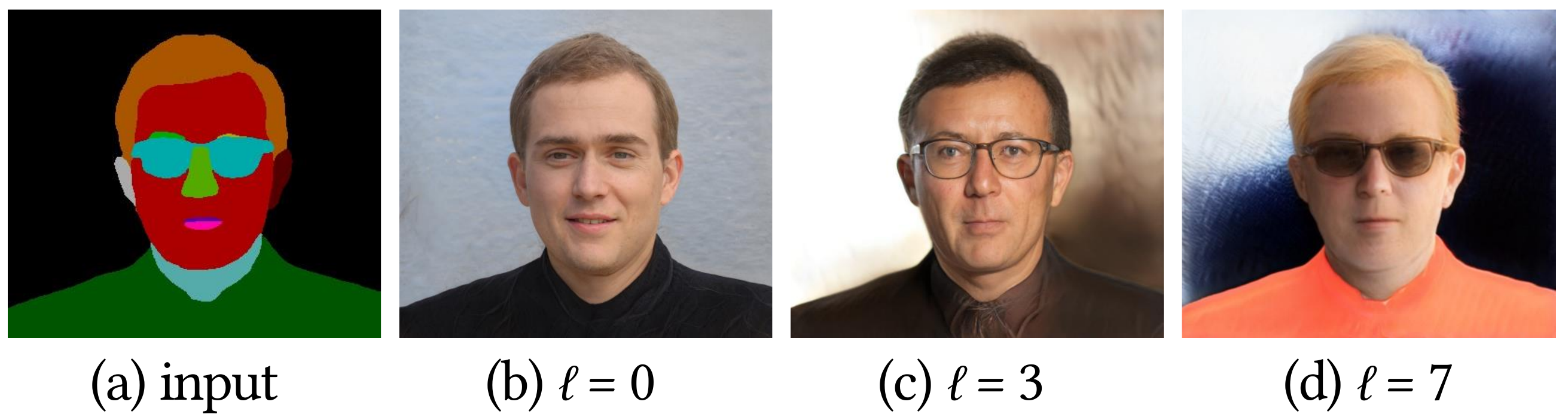}\vspace{-1.5mm}
\caption{\textbf{Effect of skip connections}.}
\label{fig:abla_ell}\vspace{1.5mm}
\includegraphics[width=\linewidth]{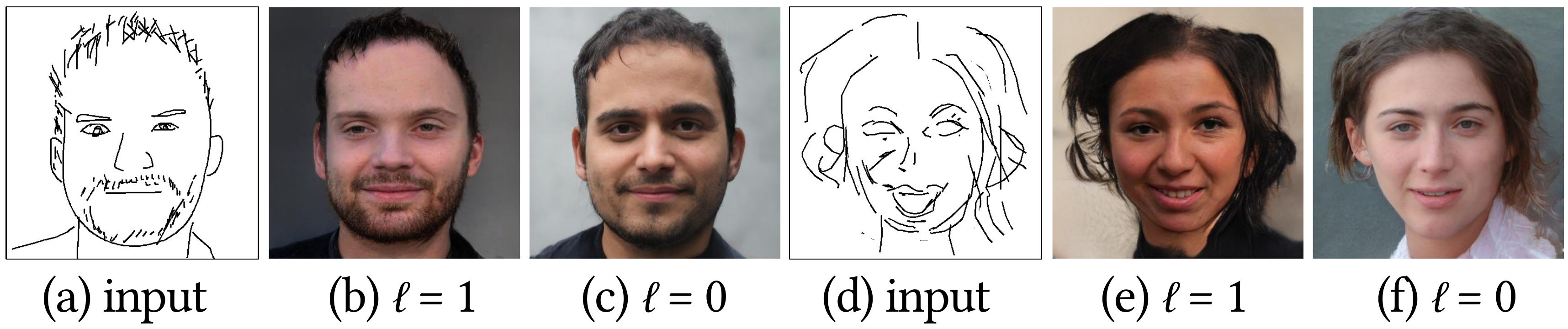}\vspace{-1.5mm}
\caption{\textbf{Performance on low-quality sketches}.}
\label{fig:abla_ell2}\vspace{2mm}
\includegraphics[width=\linewidth]{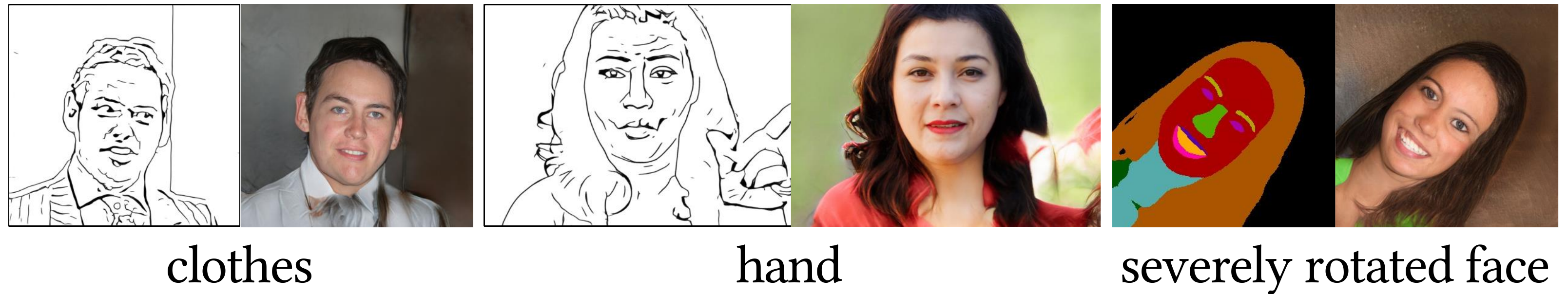}\vspace{-2mm}
\caption{\textbf{Limitations}.}
\label{fig:limitation}\vspace{-1.2em}
\end{figure}

In Fig.~\ref{fig:abla_ell}, we study the effect of skip connection. Without it ($\ell=0$), the glasses cannot be generated. Skip connection provides mid-layer features to preserve the details of the input. However, too many extra features will alter the feature distribution of StyleGAN, \eg, always generating sunglasses as the input has no segmentation of eyes. Thus, we use $\ell=3$ to balance between input-output consistency and diversity.
Inversely, we can use a small $\ell$ to enhance the model robustness to low-quality inputs.
For example, we can generate more realistic faces with $\ell=0$ on DeepFaceDrawing low-quality sketches~\cite{chen2020deepfacedrawing} as in Fig.~\ref{fig:abla_ell2}.

\subsection{Limitations}

First, our framework currently relies on an inefficient optimization process for precise reconstruction. While future work can explore more efficient inversion methods (\eg, iterative residue prediction and hyper networks), it is important to note that this paper focuses on overcoming the fixed-crop limitation of StyleGAN, rather than the specific topic of GAN inversion.
Second, StyleGANEX is limited by the feature representation of StyleGAN. While it shows great potential in handling normal FoV face images, out-of-distribution features such as complex clothing, human bodies and faces with large rotation angles may not be well handled as in Fig.~\ref{fig:limitation}. Similarly, StyleGANEX, like StyleGAN, focuses on face manipulation and may not support out-of-distribution semantical editing of non-facial regions.
Third, StyleGANEX may inherit the model bias of StyleGAN. Applying it to tasks with severe data imbalance might lead to unsatisfactory results on under-represented data.

\section{Conclusion}

We have presented an effective approach to refactor StyleGAN to overcome its fixed-crop limitation while retaining its style control abilities. The refactored model, called StyleGANEX, fully inherits the parameters of the pre-trained StyleGAN without retraining, and is thus fully compatible with the generative space of StyleGAN. We further introduced a StyleGANEX encoder to project normal FoV face images to the joint $W^+$\textendash$F$ space of StyleGANEX for real face inversion and manipulation. Our approach offers an effective solution to the general issue of StyleGAN and extends its capability beyond fixed-resolution data.


\noindent
\textbf{Acknowledgments.} This study is supported under the RIE2020 Industry Alignment Fund Industry Collaboration Projects (IAF-ICP) Funding Initiative, as well as cash and in-kind contribution from the industry partner(s). It is also supported by MOE AcRF Tier 1 (2021-T1-001-088) and NTU NAP grant.

{\small
\bibliographystyle{ieee_fullname}
\bibliography{egbib}

\begin{thebibliography}{10}\itemsep=-1pt

\bibitem{abdal2019image2stylegan}
Rameen Abdal, Yipeng Qin, and Peter Wonka.
\newblock Image2stylegan: How to embed images into the stylegan latent space?
\newblock In {\em Proc.~Int'l Conf.~Computer Vision}, pages 4432--4441, 2019.

\bibitem{alaluf2021restyle}
Yuval Alaluf, Or Patashnik, and Daniel Cohen-Or.
\newblock {Restyle}: A residual-based stylegan encoder via iterative
  refinement.
\newblock In {\em Proc.~Int'l Conf.~Computer Vision}, pages 6711--6720, 2021.

\bibitem{alaluf2022times}
Yuval Alaluf, Or Patashnik, Zongze Wu, Asif Zamir, Eli Shechtman, Dani
  Lischinski, and Daniel Cohen-Or.
\newblock Third time's the charm? image and video editing with stylegan3, 2022.

\bibitem{alaluf2022hyperstyle}
Yuval Alaluf, Omer Tov, Ron Mokady, Rinon Gal, and Amit Bermano.
\newblock Hyperstyle: Stylegan inversion with hypernetworks for real image
  editing.
\newblock In {\em Proc.~IEEE Int'l Conf.~Computer Vision and Pattern
  Recognition}, pages 18511--18521, 2022.

\bibitem{chan2020glean}
Kelvin~CK Chan, Xintao Wang, Xiangyu Xu, Jinwei Gu, and Chen~Change Loy.
\newblock Glean: Generative latent bank for large-factor image
  super-resolution.
\newblock In {\em Proc.~IEEE Int'l Conf.~Computer Vision and Pattern
  Recognition}, 2021.

\bibitem{chen2020deepfacedrawing}
Shu-Yu Chen, Wanchao Su, Lin Gao, Shihong Xia, and Hongbo Fu.
\newblock Deepfacedrawing: Deep generation of face images from sketches.
\newblock {\em {ACM} Transactions on Graphics}, 39(4):72--1, 2020.

\bibitem{deng2019arcface}
Jiankang Deng, Jia Guo, Niannan Xue, and Stefanos Zafeiriou.
\newblock Arcface: Additive angular margin loss for deep face recognition.
\newblock In {\em Proc.~IEEE Int'l Conf.~Computer Vision and Pattern
  Recognition}, pages 4690--4699, 2019.

\bibitem{dinh2022hyperinverter}
Tan~M Dinh, Anh~Tuan Tran, Rang Nguyen, and Binh-Son Hua.
\newblock Hyperinverter: Improving stylegan inversion via hypernetwork.
\newblock In {\em Proc.~IEEE Int'l Conf.~Computer Vision and Pattern
  Recognition}, pages 11389--11398, 2022.

\bibitem{gal2022stylegan}
Rinon Gal, Or Patashnik, Haggai Maron, Amit~H Bermano, Gal Chechik, and Daniel
  Cohen-Or.
\newblock Stylegan-nada: Clip-guided domain adaptation of image generators.
\newblock {\em ACM Transactions on Graphics (TOG)}, 41(4):1--13, 2022.

\bibitem{harkonen2020ganspace}
Erik H{\"a}rk{\"o}nen, Aaron Hertzman, Jaakko Lehtinen, and Sylvain Paris.
\newblock Ganspace: Discovering interpretable gan controls.
\newblock In {\em Advances in Neural Information Processing Systems}, 2020.

\bibitem{he2016deep}
Kaiming He, Xiangyu Zhang, Shaoqing Ren, and Jian Sun.
\newblock Deep residual learning for image recognition.
\newblock In {\em Proc.~IEEE Int'l Conf.~Computer Vision and Pattern
  Recognition}, pages 770--778, 2016.

\bibitem{jiang2020tsit}
Liming Jiang, Changxu Zhang, Mingyang Huang, Chunxiao Liu, Jianping Shi, and
  Chen~Change Loy.
\newblock Tsit: A simple and versatile framework for image-to-image
  translation.
\newblock In {\em Proc.~European Conf.~Computer Vision}, pages 206--222.
  Springer, 2020.

\bibitem{jiang2021talkedit}
Yuming Jiang, Ziqi Huang, Xingang Pan, Chen~Change Loy, and Ziwei Liu.
\newblock Talk-to-edit: Fine-grained facial editing via dialog.
\newblock In {\em Proc.~Int'l Conf.~Computer Vision}, 2021.

\bibitem{kang2021gan}
Kyoungkook Kang, Seongtae Kim, and Sunghyun Cho.
\newblock Gan inversion for out-of-range images with geometric transformations.
\newblock In {\em Proc.~Int'l Conf.~Computer Vision}, pages 13941--13949, 2021.

\bibitem{karras2020training}
Tero Karras, Miika Aittala, Janne Hellsten, Samuli Laine, Jaakko Lehtinen, and
  Timo Aila.
\newblock Training generative adversarial networks with limited data.
\newblock In {\em Advances in Neural Information Processing Systems}, 2020.

\bibitem{karras2021alias}
Tero Karras, Miika Aittala, Samuli Laine, Erik H{\"a}rk{\"o}nen, Janne
  Hellsten, Jaakko Lehtinen, and Timo Aila.
\newblock Alias-free generative adversarial networks.
\newblock {\em Advances in Neural Information Processing Systems}, 34, 2021.

\bibitem{karras2019style}
Tero Karras, Samuli Laine, and Timo Aila.
\newblock A style-based generator architecture for generative adversarial
  networks.
\newblock In {\em Proc.~IEEE Int'l Conf.~Computer Vision and Pattern
  Recognition}, pages 4401--4410, 2019.

\bibitem{karras2020analyzing}
Tero Karras, Samuli Laine, Miika Aittala, Janne Hellsten, Jaakko Lehtinen, and
  Timo Aila.
\newblock Analyzing and improving the image quality of stylegan.
\newblock In {\em Proc.~IEEE Int'l Conf.~Computer Vision and Pattern
  Recognition}, pages 8110--8119, 2020.

\bibitem{liu2022deepfacevideoediting}
Feng-Lin Liu, Shu-Yu Chen, Yukun Lai, Chunpeng Li, Yue-Ren Jiang, Hongbo Fu,
  and Lin Gao.
\newblock Deepfacevideoediting: Sketch-based deep editing of face videos.
\newblock {\em {ACM} Transactions on Graphics}, 41(4):167, 2022.

\bibitem{parmar2022spatially}
Gaurav Parmar, Yijun Li, Jingwan Lu, Richard Zhang, Jun-Yan Zhu, and
  Krishna~Kumar Singh.
\newblock Spatially-adaptive multilayer selection for gan inversion and
  editing.
\newblock In {\em Proc.~IEEE Int'l Conf.~Computer Vision and Pattern
  Recognition}, pages 11399--11409, 2022.

\bibitem{patashnik2021styleclip}
Or Patashnik, Zongze Wu, Eli Shechtman, Daniel Cohen-Or, and Dani Lischinski.
\newblock Styleclip: Text-driven manipulation of stylegan imagery.
\newblock In {\em Proc.~IEEE Int'l Conf.~Computer Vision and Pattern
  Recognition}, pages 2085--2094, October 2021.

\bibitem{pinkney2020resolution}
Justin~NM Pinkney and Doron Adler.
\newblock Resolution dependent gan interpolation for controllable image
  synthesis between domains.
\newblock {\em arXiv preprint arXiv:2010.05334}, 2020.

\bibitem{richardson2020encoding}
Elad Richardson, Yuval Alaluf, Or Patashnik, Yotam Nitzan, Yaniv Azar, Stav
  Shapiro, and Daniel Cohen-Or.
\newblock Encoding in style: a stylegan encoder for image-to-image translation.
\newblock In {\em Proc.~IEEE Int'l Conf.~Computer Vision and Pattern
  Recognition}, 2021.

\bibitem{roich2021pivotal}
Daniel Roich, Ron Mokady, Amit~H Bermano, and Daniel Cohen-Or.
\newblock Pivotal tuning for latent-based editing of real images.
\newblock {\em {ACM} Transactions on Graphics}, 2022.

\bibitem{roessler2019faceforensicspp}
Andreas R\"ossler, Davide Cozzolino, Luisa Verdoliva, Christian Riess, Justus
  Thies, and Matthias Nie{\ss}ner.
\newblock Face{F}orensics++: Learning to detect manipulated facial images.
\newblock In {\em Proc.~Int'l Conf.~Computer Vision}, 2019.

\bibitem{shen2020interpreting}
Yujun Shen, Jinjin Gu, Xiaoou Tang, and Bolei Zhou.
\newblock Interpreting the latent space of gans for semantic face editing.
\newblock In {\em Proc.~IEEE Int'l Conf.~Computer Vision and Pattern
  Recognition}, pages 9243--9252, 2020.

\bibitem{shen2020interfacegan}
Yujun Shen, Ceyuan Yang, Xiaoou Tang, and Bolei Zhou.
\newblock Interfacegan: Interpreting the disentangled face representation
  learned by gans.
\newblock {\em {IEEE} Transactions on Pattern Analysis and Machine
  Intelligence}, 2020.

\bibitem{shen2021closed}
Yujun Shen and Bolei Zhou.
\newblock Closed-form factorization of latent semantics in gans.
\newblock In {\em Proc.~IEEE Int'l Conf.~Computer Vision and Pattern
  Recognition}, pages 1532--1540, 2021.

\bibitem{tewari2020pie}
Ayush Tewari, Mohamed Elgharib, Florian Bernard, Hans-Peter Seidel, Patrick
  P{\'e}rez, Michael Zollh{\"o}fer, and Christian Theobalt.
\newblock Pie: Portrait image embedding for semantic control.
\newblock {\em {ACM} Transactions on Graphics}, 39(6):1--14, 2020.

\bibitem{tov2021designing}
Omer Tov, Yuval Alaluf, Yotam Nitzan, Or Patashnik, and Daniel Cohen-Or.
\newblock Designing an encoder for stylegan image manipulation.
\newblock {\em {ACM} Transactions on Graphics}, 40(4):1--14, 2021.

\bibitem{tzaban2022stitch}
Rotem Tzaban, Ron Mokady, Rinon Gal, Amit Bermano, and Daniel Cohen-Or.
\newblock Stitch it in time: Gan-based facial editing of real videos.
\newblock In {\em SIGGRAPH Asia}, pages 1--9, 2022.

\bibitem{unterthiner2018towards}
Thomas Unterthiner, Sjoerd van Steenkiste, Karol Kurach, Raphael Marinier,
  Marcin Michalski, and Sylvain Gelly.
\newblock Towards accurate generative models of video: A new metric \&
  challenges.
\newblock {\em arXiv preprint arXiv:1812.01717}, 2018.

\bibitem{viazovetskyi2020stylegan2}
Yuri Viazovetskyi, Vladimir Ivashkin, and Evgeny Kashin.
\newblock Stylegan2 distillation for feed-forward image manipulation.
\newblock In {\em Proc.~European Conf.~Computer Vision}, pages 170--186.
  Springer, 2020.

\bibitem{wang2022high}
Tengfei Wang, Yong Zhang, Yanbo Fan, Jue Wang, and Qifeng Chen.
\newblock High-fidelity gan inversion for image attribute editing.
\newblock In {\em Proc.~IEEE Int'l Conf.~Computer Vision and Pattern
  Recognition}, pages 11379--11388, 2022.

\bibitem{Wang2017High}
Ting~Chun Wang, Ming~Yu Liu, Jun~Yan Zhu, Andrew Tao, Jan Kautz, and Bryan
  Catanzaro.
\newblock High-resolution image synthesis and semantic manipulation with
  conditional gans.
\newblock In {\em Proc.~IEEE Int'l Conf.~Computer Vision and Pattern
  Recognition}, 2018.

\bibitem{wang2021real}
Xintao Wang, Liangbin Xie, Chao Dong, and Ying Shan.
\newblock Real-esrgan: Training real-world blind super-resolution with pure
  synthetic data.
\newblock In {\em Proc.~Int'l Conf.~Computer Vision}, pages 1905--1914, 2021.

\bibitem{wu2021stylealign}
Zongze Wu, Yotam Nitzan, Eli Shechtman, and Dani Lischinski.
\newblock {StyleAlign}: Analysis and applications of aligned stylegan models.
\newblock {\em arXiv preprint arXiv:2110.11323}, 2021.

\bibitem{yang2022Vtoonify}
Shuai Yang, Liming Jiang, Ziwei Liu, and Chen~Change Loy.
\newblock Vtoonify: Controllable high-resolution portrait video style transfer.
\newblock {\em {ACM} Transactions on Graphics}, 41(6):1--15, 2022.

\bibitem{yin2022styleheat}
Fei Yin, Yong Zhang, Xiaodong Cun, Mingdeng Cao, Yanbo Fan, Xuan Wang, Qingyan
  Bai, Baoyuan Wu, Jue Wang, and Yujiu Yang.
\newblock Styleheat: One-shot high-resolution editable talking face generation
  via pre-trained stylegan.
\newblock In {\em Proc.~European Conf.~Computer Vision}, pages 85--101.
  Springer, 2022.

\bibitem{Yu-ECCV-BiSeNet-2018}
Changqian Yu, Jingbo Wang, Chao Peng, Changxin Gao, Gang Yu, and Nong Sang.
\newblock Bisenet: Bilateral segmentation network for real-time semantic
  segmentation.
\newblock In {\em Proc.~European Conf.~Computer Vision}, pages 334--349.
  Springer, 2018.

\bibitem{zhu2021low}
Jiapeng Zhu, Ruili Feng, Yujun Shen, Deli Zhao, Zheng-Jun Zha, Jingren Zhou,
  and Qifeng Chen.
\newblock Low-rank subspaces in gans.
\newblock In {\em Advances in Neural Information Processing Systems},
  volume~34, 2021.

\bibitem{zhu2020domain}
Jiapeng Zhu, Yujun Shen, Deli Zhao, and Bolei Zhou.
\newblock In-domain gan inversion for real image editing.
\newblock In {\em Proc.~European Conf.~Computer Vision}, pages 592--608.
  Springer, 2020.

\bibitem{zhu2021barbershop}
Peihao Zhu, Rameen Abdal, John Femiani, and Peter Wonka.
\newblock Barbershop: Gan-based image compositing using segmentation masks.
\newblock {\em {ACM} Transactions on Graphics}, 40(6):1--13, 2021.

\end{thebibliography}
}

\clearpage

\appendix

\section{Appendix: Implementation Details}

\subsection{Network Architecture}

PSp~\cite{richardson2020encoding} has multi-scale intermediate layers, where 1-3 layers are for $128\times128$ features, the middle 4-7 layers are for $64\times64$ features and the subsequent 8-21 layers are for $32\times32$ features.
For StyleGANEX encoder, we concatenate three features from layers 11, 16 and 21 and add a convolution layer to map the concatenated features to the first-layer input feature $f$.

For skip connections, we use the features from layers 0 (the layer before the intermediate layers), 3, 7, 11, 16, 21, 21 as the skipped features into the StyleGANEX. These seven features are skipped to the StyleGANEX layers corresponding to the resolution 256, 128, 64, 32, 16, 8, 4 of StyleGAN, respectively (corresponding to $\ell=$13, 11, 9, 7, 5, 3, 1).
The skipped feature and the StyleGANEX feature are concatenated and go through an added convolution layer to obtain the fused feature to have the same resolution and channel numbers as the original StyleGANEX feature.

\subsection{Running Time}

\noindent
\textbf{Training encoder} uses one NVIDIA Tesla V100 GPU for 100,000 iterations for all tasks except that video toonification uses 50,000 iterations. The training time is about 2 days for 100,000 iterations and 1 day for 50,000 iterations, respectively.

\noindent
\textbf{Image Inference} uses one NVIDIA Tesla V100 GPU and a batch size of $1$. All the following running time includes IO and face detection.
Inferencing on 796 testing images which is of averaged $360\times398$ size (the corresponding output image is about $1440\times1592$), the inversion of each image is about 107.11 s, where the fast feed-forward Step I takes about 0.386 s.
For other fast feed-forward tasks such as super-resolution and translation take about 0.259 s-0.545 s depending on the network architectures (\ie how many skip connection layers, whether using $T$).

\noindent
\textbf{Video Inference} uses one NVIDIA Tesla V100 GPU and a batch size of $4$. All the following running time includes IO and face detection.
Inferencing on 28 ten-second video clips, which is of averaged $338\times398$ size (the corresponding edited video is about $1352\times1592$), the video editing/toonification takes about 45 s per video.

\section{Appendix: Compatibility to StyleGAN}

\begin{figure}[t]
\centering
\includegraphics[width=\linewidth]{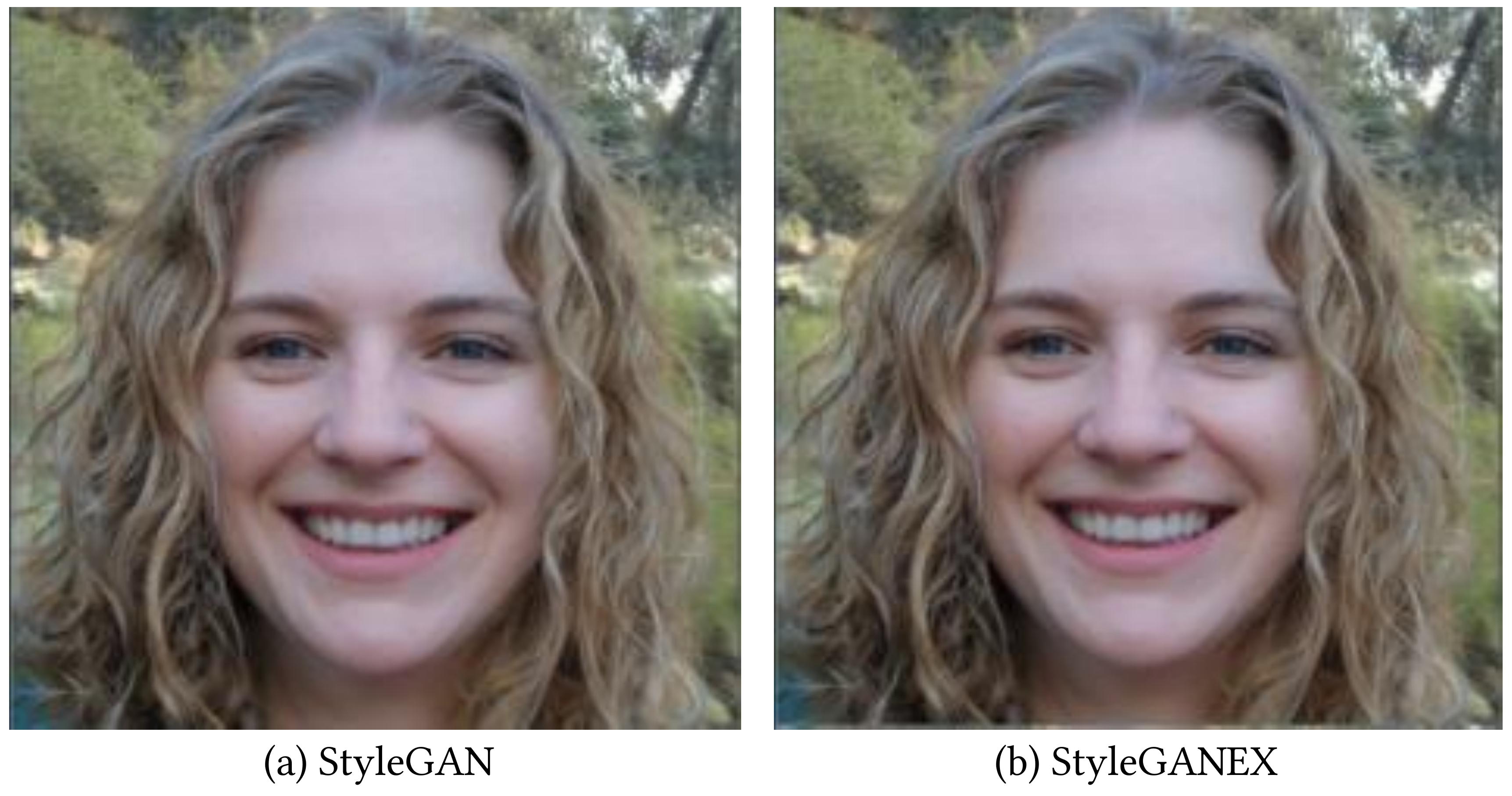}
\caption{\textbf{StyleGANEX is compatible with StyleGAN}.}
\label{fig:compatible}
\end{figure}

StyleGANEX is fully compatible with StyleGAN and can directly load a pre-trained StyleGAN model without training. In Fig.~\ref{fig:compatible}, we upsample the StyleGAN's constant input feature $f_0$ by $8\times$ with nearest neighbor interpolation to serve as the first-layer feature of StyleGANEX, StyleGANEX generates the same face image as the StyleGAN from the same latent code $w^+$. Formally, we have $G(f_{0\uparrow}, w^+)=G_0(w^+)$, where $G$ and $G_0$ are  StyleGANEX and StyleGAN, respectively. $f_{0\uparrow}$ is the $8\times$  upsampled $f_0$ with nearest neighbor interpolation.

\section{Appendix: Full Image Stylization}

\begin{figure*}[h]
\centering
\includegraphics[width=\linewidth]{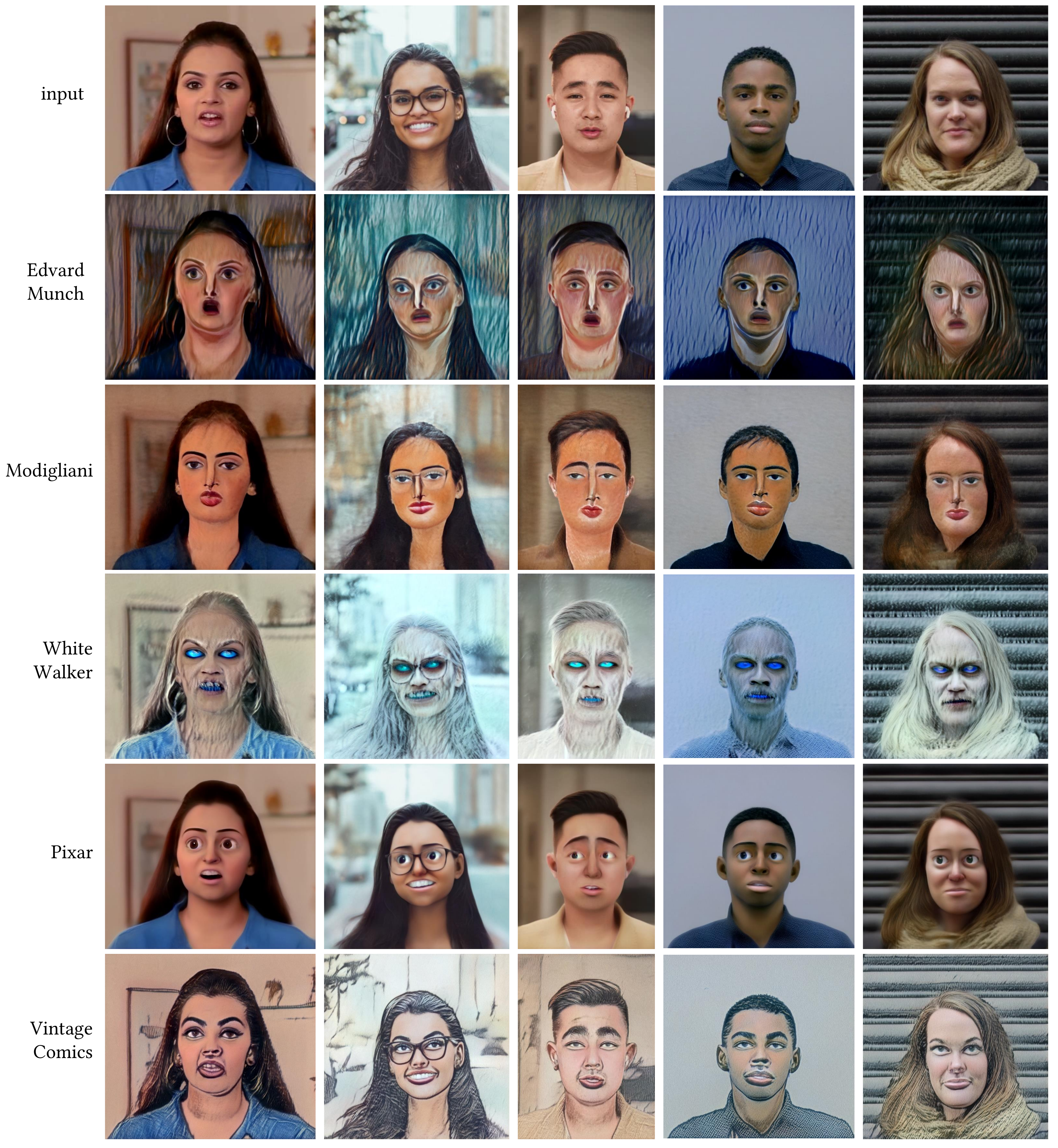}
\caption{\textbf{Full image stylization results}.}
\label{fig:domain_ada}
\end{figure*}

After StyleGANEX inversion, we can achieve full image style transfer by loading a new domain adapted StyleGAN model.
Here, we show results on five different StyleGAN models provided by StyleGAN-NADA~\cite{gal2022stylegan} in Fig.~\ref{fig:domain_ada}.
Remarkably, our method successfully renders the full background with the target style, which cannot be simply achieved by cropping, editing and pasting.

\section{Appendix: Supplementary Comparison}

\subsection{Qualitative Evaluation}

\noindent
\textbf{Normal FoV face inversion}
Figure~\ref{fig:inversion1}  compares with pSp~\cite{richardson2020encoding} and HyperStyle~\cite{alaluf2022hyperstyle} on normal FoV face inversion. Our encoder surpasses the baselines in handling the complete scenes. With Step-II optimization, our method can further precisely reconstruct the details.

\begin{table*} [t]
\caption{\textbf{Qualitative evaluation of video editing}. Best scores are marked in bold.}\vspace{-2mm}
\label{tb:editing}
\centering
\begin{tabular*}{0.9\textwidth}{@{\extracolsep{\fill}}l|c|c|c|c|c|c|c|c|c}
\toprule
Task & \multicolumn{3}{c|}{hair color editing} & \multicolumn{3}{c|}{age editing} & \multicolumn{3}{c}{average}\\
\midrule
Metric & ID-c$\downarrow$ & ID-m$\downarrow$ & FVD$\downarrow$ & ID-c$\downarrow$ & ID-m$\downarrow$ & FVD$\downarrow$ & ID-c$\downarrow$ & ID-m$\downarrow$ & FVD$\downarrow$\\
\midrule
pSp~\cite{richardson2020encoding} &  0.174 & 0.324 & 1212.72 & 0.167 & 0.326 & 1277.92 & 0.171 & 0.325 & 1245.32 \\
HyperStyle~\cite{alaluf2022hyperstyle} & 0.117 & 0.319 & 416.71  & 0.130 & 0.320 & 448.56 & 0.124 & 0.320 & 432.64 \\
ours & \textbf{0.048} & \textbf{0.312} & \textbf{186.31} & \textbf{0.055} & \textbf{0.308} & \textbf{210.38} & \textbf{0.052} & \textbf{0.310} & \textbf{198.35}  \\
\bottomrule
\end{tabular*}\vspace{-2mm}
\end{table*}

\begin{figure*}[h]
\centering
\includegraphics[width=\linewidth]{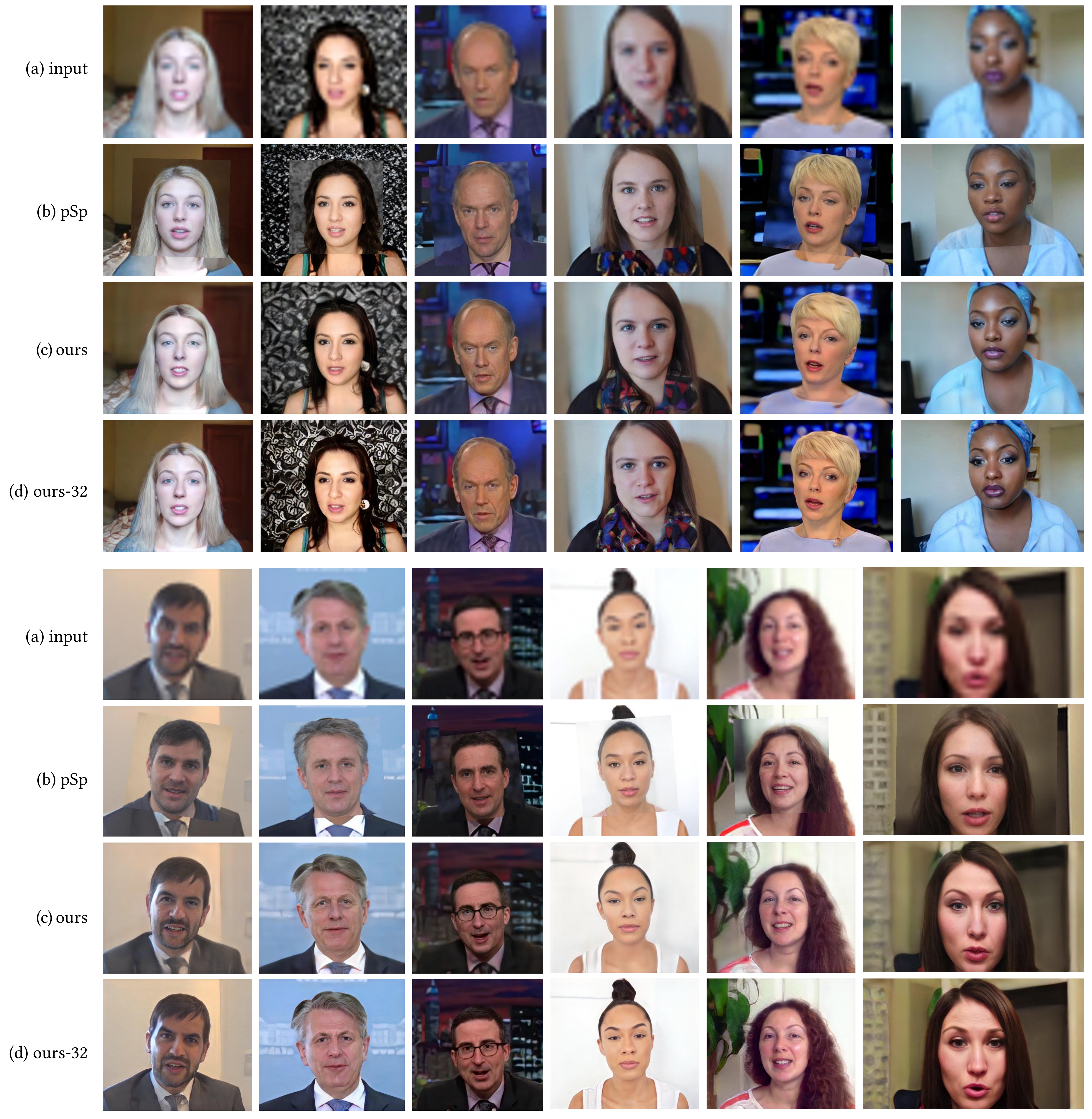}\vspace{-2mm}
\caption{\textbf{Comparison on super-resolution}.}\vspace{-2mm}
\label{fig:sr1}
\end{figure*}

\noindent
\textbf{Normal FoV face super-resolution}
We show $32\times$ super-resolution results in Fig.~\ref{fig:sr1}, where both the face and non-face regions are reasonably restored. We follow pSp to train a single mode on multiple rescaling factors ($4\sim64$) with $\ell=3$ to make a fair comparison. In pSp's results, non-face regions are super-resolved by Real-ESRGAN~\cite{wang2021real}. As in Fig.~\ref{fig:sr1}(b)(c), our method surpasses pSp in detail restoration (\eg, glasses) and uniform super-resolution without discontinuity between face and non-face regions.

\begin{figure*}[h]
\centering
\includegraphics[width=\linewidth]{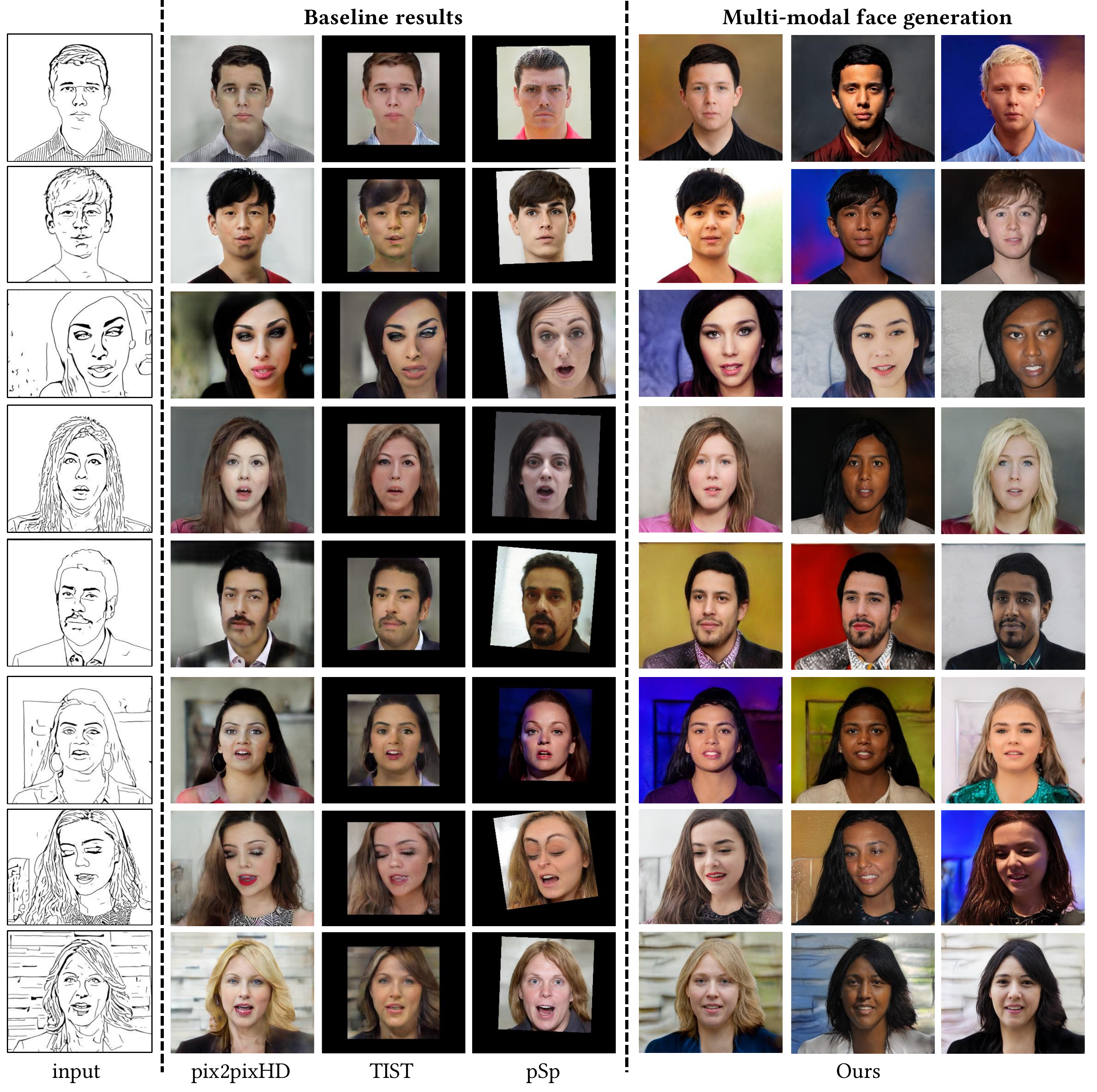}
\caption{\textbf{Comparison on sketch-to-face translation}.}
\label{fig:supp_sketch2face}
\end{figure*}

\begin{figure*}[h]
\centering
\includegraphics[width=\linewidth]{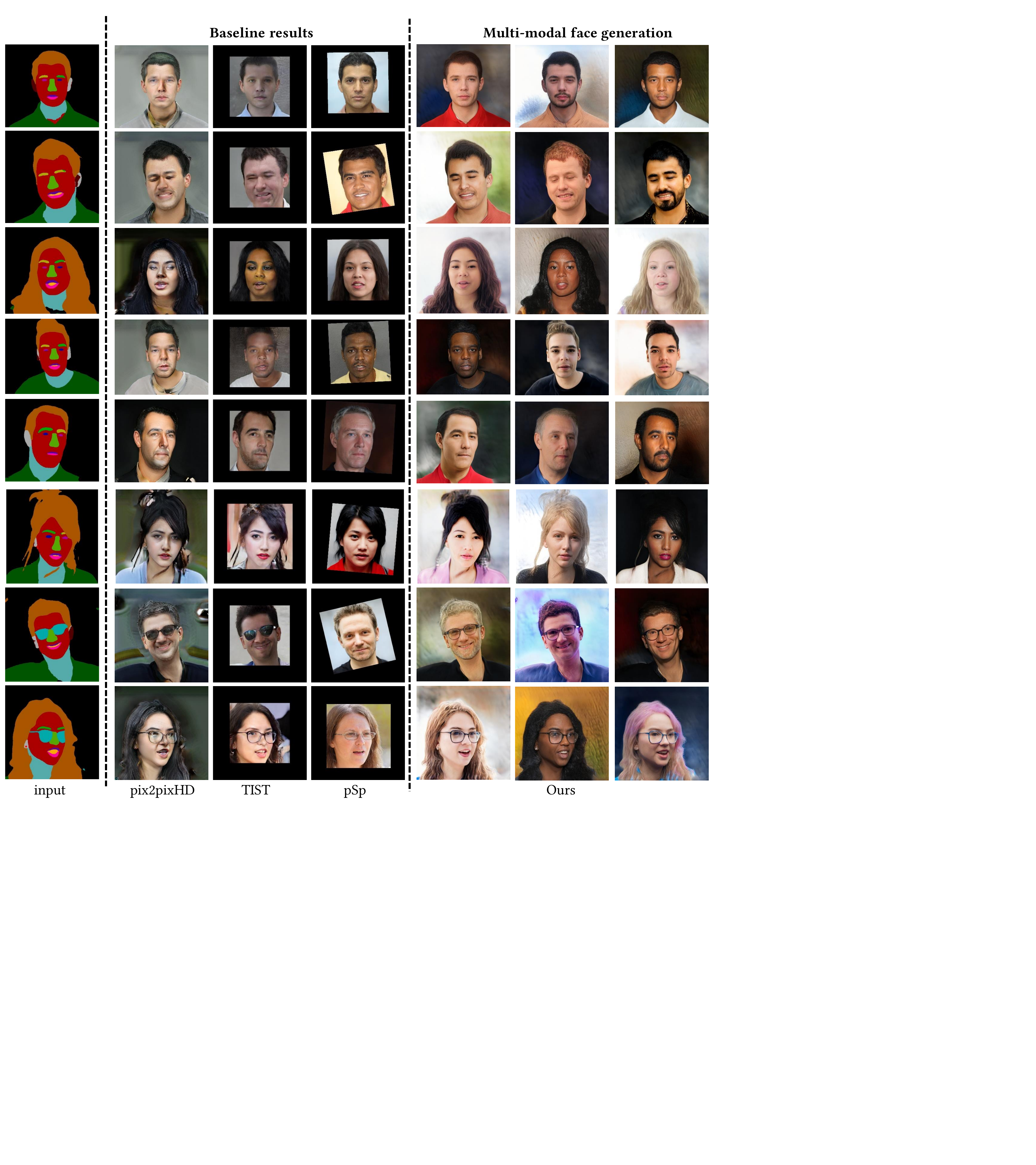}
\caption{\textbf{Comparison on mask-to-face translation}.}
\label{fig:supp_seg2face}
\end{figure*}

\noindent
\textbf{Sketch/mask-to-face translation}
We compare our method with pix2pixHD~\cite{Wang2017High}, TSIT~\cite{jiang2020tsit}, and pSp in Figs.~\ref{fig:supp_sketch2face}-\ref{fig:supp_seg2face}. Pix2pixHD's results have many artifacts and monotonous colors. TSIT' results are blurry. PSp generates realistic results, which are however less similar to the input sketch/mask. Our method can translate whole images and achieve realism and structural consistency to the inputs. Furthermore, our method supports multi-modal face generation by sampling style latent codes in the deep $11$ layers.

Compared to our method,  pix2pixHD~\cite{Wang2017High} pays more attention to keep consistency with the input sketches or masks.  For high-quality inputs, pix2pixHD sometimes has overall good translation results (\eg, the third example in Fig.~\ref{fig:supp_sketch2face}). This is why the superiority of our method is not evident in the user study (Table~\ref{tb:user_study}) of our main paper. However, for low-quality inputs, our method will be more robust than pix2pixHD. In addition, our method can use adaptive $\ell$ to meet the practical requirement (\ie, more consistency or more robust) as analyzed in Fig.~\ref{fig:abla_ell} of our main paper.

\noindent
\textbf{Video face attribute editing}
Figure~\ref{fig:supp_video_editing} compares with pSp and HyperStyle on video face attribute editing.
The results of the baselines have discontinuity near the seams. In addition, the latent code alone cannot ensure temporal consistency in videos.  The hair details randomly vary without consistency. By comparison, our method uses the first-layer feature and skipped mid-layer features to provide spatial information, which achieves more coherent results.

\begin{figure*}[h]
\centering
\includegraphics[width=\linewidth]{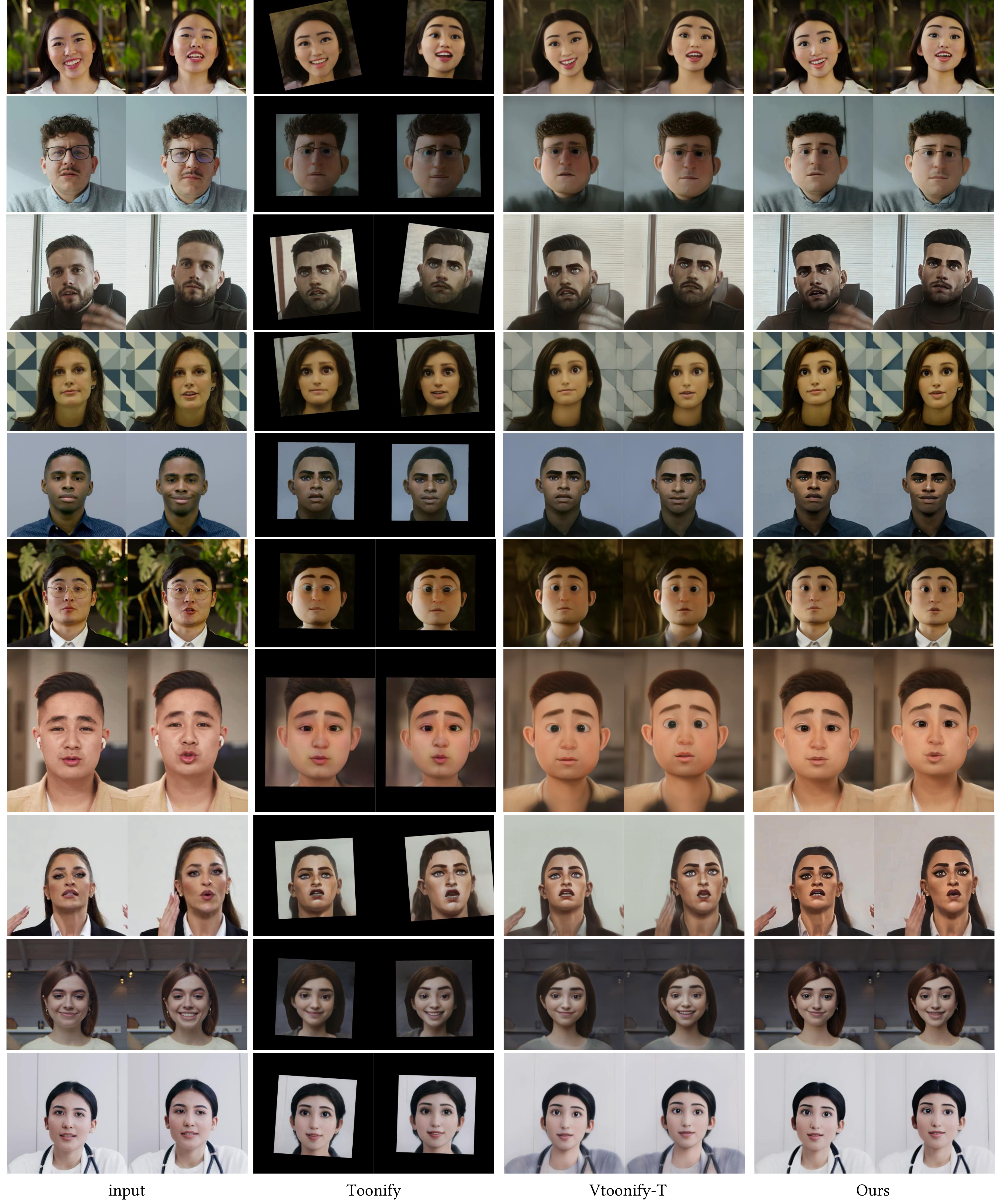}
\caption{\textbf{Comparison on video toonify}.}
\label{fig:video_toonify1}
\end{figure*}

\noindent
\textbf{Video face toonification}
Figure~\ref{fig:video_toonify1} compare with Toonify~\cite{pinkney2020resolution} and VToonify-T~\cite{yang2022Vtoonify} where our method preserves more details of the non-face region and generates sharper faces and hairs. The reason is that VToonify-T uses a fixed latent code extractor while our method trains a joint latent code and feature extractor, thus our method is more powerful for reconstructing the details. Moreover, our method retains StyleGAN's shallow layers, which helps provide key facial features to make the stylized face more vivid.

\begin{table} [t]
\caption{\textbf{Qualitative evaluation of inversion}.}\vspace{-2mm}
\label{tb:inversion}
\centering
\begin{tabular}{l|c|c|c}
\toprule
Metric & LPIPS$\downarrow$ & MAE$\downarrow$ & MSE$\downarrow$ \\
\midrule
pSp~\cite{richardson2020encoding} & 0.539 & 0.486 & 0.547  \\
HyperStyle~\cite{alaluf2022hyperstyle} & 0.518 & 0.472 & 0.542 \\
\midrule
only Step I & 0.385 & 0.130 & 0.039  \\
only Step II & 0.120 & 0.122 & 0.055  \\
ours & \textbf{0.086} & \textbf{0.057} & \textbf{0.012}  \\
\bottomrule
\end{tabular}
\end{table}

\subsection{Quantitative Evaluation}

Besides the user studies on the tasks of sketch/mask-to-face translation and video toonify in the main paper, we present more quantitative comparisons on other tasks. Since our baselines are mainly designed for cropped aligned faces, their results either have unprocessed black regions or discontinuities along the seams between the original regions.
In this case, it might be hard to find appropriate evaluation metrics from previous studies for rigorous comparison.
Therefore, the quantitative scores in this section are just for reference of the performance and we did not include them in the main paper.

\begin{figure*}[h]
\centering
\includegraphics[width=\linewidth]{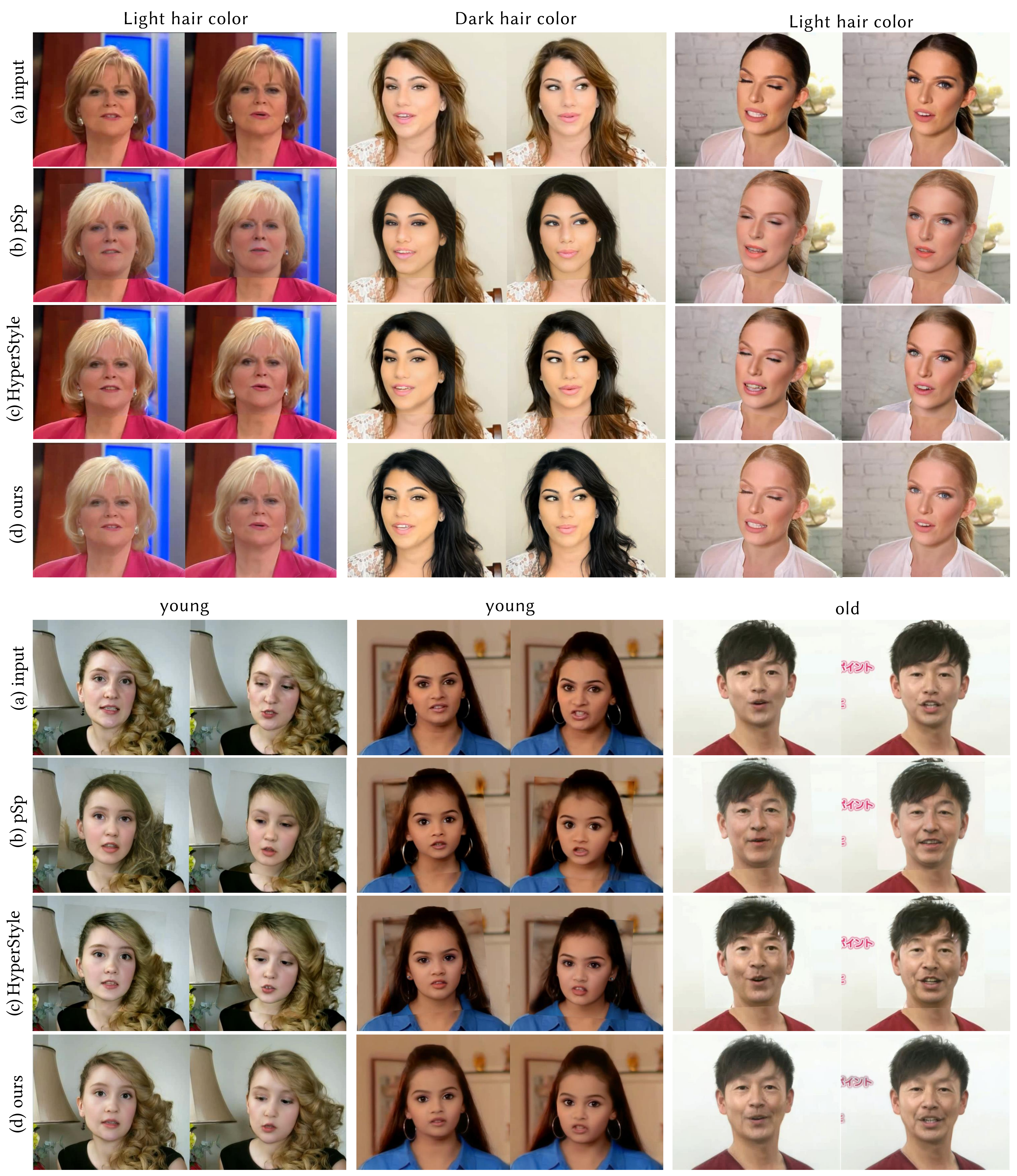}
\caption{\textbf{Comparison on video face attribute editing}.}
\label{fig:supp_video_editing}
\end{figure*}

\noindent
\textbf{Video face attribute editing}
We use 28 videos from FaceForensics++~\cite{roessler2019faceforensicspp} as a testing set to evaluate the quality of face attribute editing. Some examples are shown in Fig.~\ref{fig:supp_video_editing}.
For temporal consistency, we use ID-c and ID-m as metrics:
\begin{itemize}[itemsep=1.5pt,topsep=1pt,parsep=0pt]
  \item Identity consistency (ID-c): It measures the consistency between the edited face and the input face. We calculate the identity loss~\cite{deng2019arcface} between each edited frame and the original frame.
  \item Identity maintenance (ID-m): It measures the preservation of the identity along all edited frames. For each edited video clip, we calculate the identity loss between the generated frames and the first edited frame.
\end{itemize}
Table~\ref{tb:editing} reports the averaged ID-c and ID-m over all the video clips and our method achieves the best temporal consistency in terms of identity consistency and maintenance.

For video quality, we use frechet video distance (FVD)~\cite{unterthiner2018towards} as the evaluation metric. We resize all videos to $224\times224$ and use the first $150$ frames of each video to calculate FVD.
Table~\ref{tb:editing} reports the averaged FVD over two editing tasks, and our method obtains the highest video quality in both tasks.

\begin{table} [t]
\caption{\textbf{Qualitative evaluation of super-resolution}.}\vspace{-2mm}
\label{tb:sr}
\centering
\resizebox{\linewidth}{!}{
\begin{tabular}{l|c|c|c}
\toprule
Metric & LPIPS$\downarrow$ & MAE$\downarrow$ & PSNR$\uparrow$ \\
\midrule
pSp~\cite{richardson2020encoding} + Real-ESRGAN~\cite{wang2021real} & 0.386 & 0.105 & 21.638  \\
ours & 0.356 & 0.084 & 24.257 \\
\midrule
ours-32 & \textbf{0.304} & \textbf{0.068} & \textbf{25.617}  \\
\bottomrule
\end{tabular}}
\end{table}

\begin{figure*}[h]
\centering
\includegraphics[width=\linewidth]{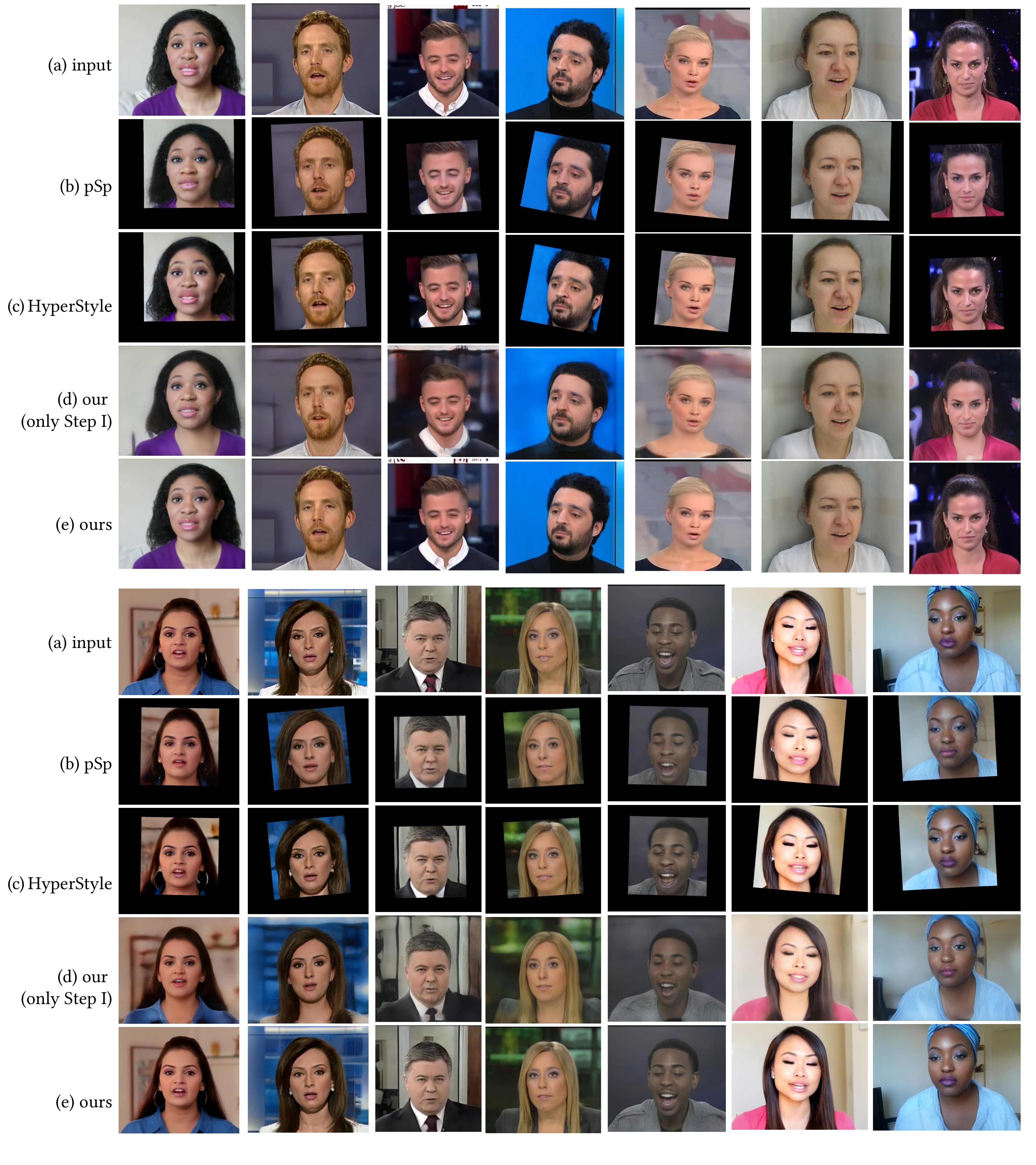}
\caption{\textbf{Comparison on normal FoV face inversion}.}
\label{fig:inversion1}
\end{figure*}

\noindent
\textbf{Normal FoV face inversion}
We use the first frame of 796 videos from FaceForensics++~\cite{roessler2019faceforensicspp} as a testing set to evaluate the quality of StyleGAN inversion. Part of the examples are shown in Fig.~\ref{fig:inversion1}.
We compare the LPIPS distance, the mean absolute error (MAE) and the mean squared error (MSE) between the reconstructed image and the input image. The results are shown in Table~\ref{tb:inversion}.
It can be seen that the unprocessed black regions in pSp and HyperStyle's results greatly harm their scores.
By comparison, our encoder (Step I) achieves better scores. Our full two-step inversion obtains the best scores.
For a fair comparison, we use 500 iterations for all optimizations in Step II.

\noindent
\textbf{Normal FoV face super-resolution}
We use the first frame of 796 videos from FaceForensics++~\cite{roessler2019faceforensicspp} as a testing set to evaluate the quality of face super-resolution. Part of the examples are shown in Fig.~\ref{fig:sr1}. PSp pays attention to the realism of the face, but lacks fidelity to the inputs. By comparison, our results are more consistent with the input faces, thus obtaining better scores in LPIPS, MAE and PSNR.

\section{Appendix: Results on Non-Facial Dataset}

\begin{figure}[htbp]
\centering
\includegraphics[width=\linewidth]{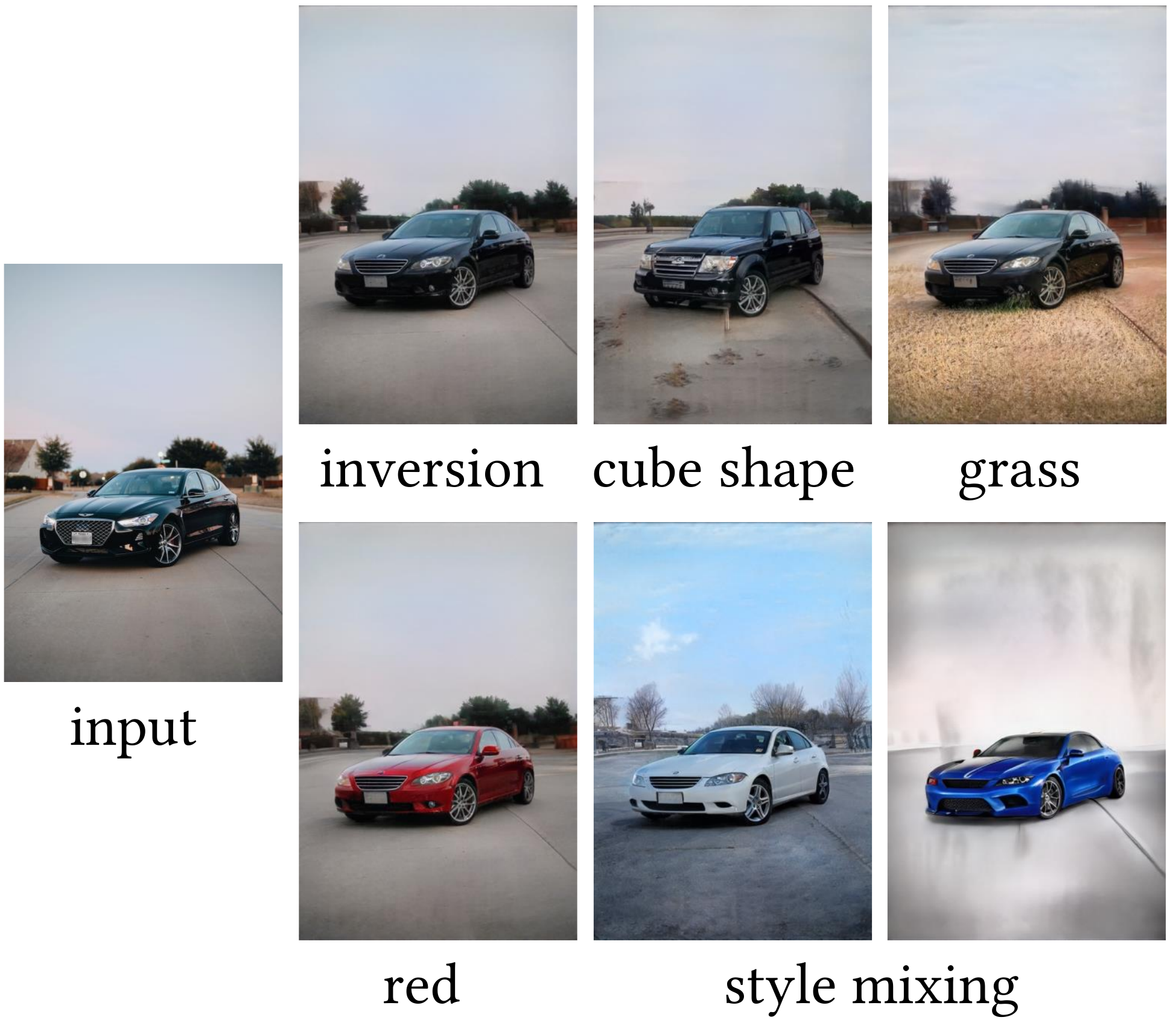}
\caption{\textbf{Performance on car}.}
\label{fig:car}
\end{figure}

 Our refactoring is domain-agnostic, thus it can be applied to other domains like cars in Fig.~\ref{fig:car}.

\end{document}